\documentclass[10pt,twocolumn,letterpaper]{article}

\usepackage{cvpr}              

\usepackage{hyperref}
\usepackage{url}
\usepackage{amsmath, amssymb}
\usepackage{graphicx}
\usepackage[table]{xcolor}
\usepackage{wrapfig}
\usepackage{booktabs}
\usepackage{multirow}
\usepackage{array}
\usepackage{makecell}
\usepackage{caption}
\usepackage{ragged2e} 
\usepackage{tabularx}       
\usepackage{subcaption}
\newcommand{\cellimg}[1]{\includegraphics[height=2.1cm,keepaspectratio]{#1}}

\newcolumntype{Q}{>{\raggedright\arraybackslash\small}p{0.22\linewidth}}
\newcolumntype{C}{>{\centering\arraybackslash}m{0.145\linewidth}}

\usepackage{amsfonts}       
\usepackage[export]{adjustbox}
\usepackage{booktabs}       

\usepackage{minitoc}

\def\paperID{*****} 
\def\confName{CVPR}
\def\confYear{2026}

\title{Understanding and Harnessing Sparsity in Unified Multimodal Models
} 

\author{Shwai He\textsuperscript{\rm 1,2
,\thanks{Work done while the first author was an intern at ByteDance Seed.}
}
\space\space\space\space
Chaorui Deng\textsuperscript{\rm 1}\space\space\space\space
Ang Li\textsuperscript{\rm 2}\space\space\space
Shen Yan\textsuperscript{\rm 1}
\space\space\space\space \\
    \textsuperscript{\rm 1}ByteDance Seed \space\space
    \textsuperscript{\rm 2}University of Maryland, College Park \\
    {\tt\small shwai.he@bytedance.com, sheny@bytedance.com}}

\begin{document}
\maketitle

\begin{abstract}
Large multimodal models have achieved remarkable progress in both understanding and generation. 
Recent efforts pursue unified multimodal models that integrate heterogeneous components to support both capabilities within a single framework. 
However, such unification introduces inference inefficiencies, e.g., specific tasks or samples may not require the full knowledge or capacity of the unified model. 
Yet, a systematic understanding of how these inefficiencies manifest across different components remains limited.
In this work, we first conduct a systematic analysis of unified multimodal model components using training-free pruning as a probing methodology, considering both depth pruning and width reduction. 
Our study reveals that the understanding component exhibits notable compressibility in both understanding and generation tasks, which is more pronounced in the latter. In contrast, the generation components are highly sensitive to compression, with performance deteriorating sharply even under moderate compression ratios. 
To address this limitation, we propose the Mixture-of-Experts (MoE) Adaptation, inspired by the dynamic activation patterns observed across different samples. 
This approach partitions the generation module into multiple experts and enables sparse activation to restore generation quality. 
We validate the effectiveness of sparse activation through expert-frozen tuning and further demonstrate that a fully trainable adaptation delivers additional gains. 
As a result, the adapted BAGEL model achieves performance comparable to the full model while activating only about half of its parameters. 
The code is released at 
\href{https://github.com/Shwai-He/SparseUnifiedModel}{this link}. 
\end{abstract}

\section{Introduction}

Large-scale multimodal models have recently achieved remarkable progress in both multimodal understanding \citep{liu2023visual, li2023blip, dai2023instructblip, lu2024deepseekvl, lu2023chameleon} and generation \citep{pmlr-v139-ramesh21a, saharia2022photorealistic, peebles2023scalablediffusionmodelstransformers}. Traditionally, these two tasks were studied in isolation, leading to distinct research trajectories and model families: \textit{understanding models} for vision–language reasoning with textual outputs, and \textit{generative models} designed for image synthesis. While effective for task-specific purposes, this separation stands in contrast to the broader pursuit \citep{wei2022emergent, bubeck2023sparksartificialgeneralintelligence}, which envisions a unified model capable of both understanding and generating across modalities. 
Recent research has increasingly explored unified multimodal models \citep{wu2024janus, chen2025janus, deng2025emergingpropertiesunifiedmultimodal, liang2025mixtureoftransformers, ai2025mingomniunifiedmultimodalmodel}, which integrate heterogeneous components such as vision encoders \citep{dosovitskiy2021an}, language backbones \citep{grattafiori2024llama3herdmodels, qwen2}, and image decoders \citep{peebles2023scalablediffusionmodelstransformers, ai2025mingomniunifiedmultimodalmodel}. These unified architectures can seamlessly support both understanding and generation within a single system, marking a promising step toward general-purpose multimodal intelligence. 

However, this unification comes at a substantial \textbf{\textit{cost in efficiency}}, as many tasks or input samples do not require the full knowledge or capacity of the unified model, but rather rely on a much slimmer or more sparsely activated sub-architecture \cite{Frankle2018TheLT, xu-etal-2021-raise}. 
In addition, we identify three complementary observations that shed light on the slimness and sparsity manifested in unified multimodal models:
(1) \textbf{\textit{Component-wise redundancy}}: 
the understanding and generation components follow distinct computation patterns and serve different functional roles, leading to different levels of redundancy. 
(2) \textbf{\textit{Task-specific activation}}: 
task-specific inputs tend to activate only a limited subset of parameters during inference, leaving much of the shared capacity underutilized \cite{kudugunta-etal-2021-beyond-distillation, Sarkar2023EdgeMoEMM}. 
(3) \textbf{\textit{Input variability}}:
even within the same task, different input queries can activate different portions of the model, further complicating efficient compute allocation \cite{liu2024trainingfreeactivationsparsitylarge, luo2025sparsing}.
    
Motivated by these factors, We systematically analyze the components of unified multimodal models and examine how they allocate and utilize parameters across tasks and inputs. 
Specifically, we first adopt \textit{\textbf{training-free pruning}} as a probing methodology \citep{gromov2025the, men2024shortgptlayerslargelanguage, he2024matterstransformersattentionneeded, ma2023llmpruner, xia2023sheared}, as it enables us to infer structural importance without retraining by removing structures and observing the resulting performance change.
We begin by analyzing the understanding components, which serve as shared modules that process inputs from multiple modalities and produce language representations or embeddings that guide the generation components. Our results demonstrate that the understanding components are highly compressible in multimodal generation tasks but less so in understanding tasks. 
Furthermore, we observe clear task-specific activation patterns: understanding and generation tasks predominantly activate different model partitions, underscoring the necessity of dynamic activation for different testing tasks. 

In contrast to understanding components, generation components (e.g., image generators) are far less tolerant to static compression, where we find the activation patterns vary significantly across samples and timesteps. Consequently, applying static pruning leads to a drastic drop in the generated image quality due to the inability to accommodate such dynamic activation shifts. 
To address this issue, 
we propose a \textbf{\textit{Mixture-of-Experts (MoE) Adaptation}}, where MLP neurons are partitioned into experts, allowing the model to selectively activate subsets of neurons to ensure efficient dynamic activation. 
To align the model with sparsely routed inference, we first introduce expert-frozen tuning, which freezes all experts and updates only the remaining parameters. 
Building on this initialization, we further perform fully 
end-to-end 
MoE training, yielding additional improvements, where the adapted BAGEL model \citep{deng2025emergingpropertiesunifiedmultimodal} matches the performance of the full model while activating only about half of neurons. Our contributions are threefold: 

\begin{itemize}
    \item Our study demonstrates that the understanding component in unified multimodal models has high structural redundancy and can be compressed even more aggressively for generation tasks.
    
    \item We introduce a training-free neuron partition analysis that uncovers task-specific activation patterns and the corresponding redundancy within the understanding component, offering insights into task-aware compression. 
    
    \item To cope with the high compression sensitivity of the generation component, we design a tailored MoE Adaptation that preserves generation quality while activating only about half of its parameters. 
\end{itemize}
\section{Related Works}

\paragraph{\textbf{Multimodal Models for Understanding and Generation}} 
Early multimodal research typically treated understanding and generation as separate tasks, leading to two distinct architectural paradigms \citep{li2023blip, dai2023instructblip, zhu2023minigpt}. 
On the one hand, multimodal large language models (MLLMs) extend language models to handle input tokens from multiple modalities, such as LLaVA~\citep{liu2023visual}, which augments the LLaMA backbone~\citep{touvron2023llamaopenefficientfoundation} with visual tokens for vision–language understanding. 
On the other hand, multimodal generative models employ dedicated generators that can synthesize high-fidelity visual outputs~\citep{Ramesh2022HierarchicalTI, Saharia2022PhotorealisticTD, Chang2023MuseTG}. For instance, recent diffusion-based approaches, including Diffusion Transformers (DiT)~\citep{peebles2023scalablediffusionmodelstransformers}, further show that iterative denoising and latent refinement can be effectively leveraged to convert noise into images guided by natural language. 
More recently, several works aim to unify understanding and generation within a single framework \citep{wu2024janus, chen2025janus, zhang2025unified}. BAGEL \citep{deng2025emergingpropertiesunifiedmultimodal} employs 
a mixture-of-transformers design~\citep{liang2025mixtureoftransformers} to separate understanding and generation modules, while Ming-Omni \citep{ai2025mingomniunifiedmultimodalmodel} uses a mixture-of-experts backbone with multimodal understanding and modality-specific decoders for generation. 
Although these unified models demonstrate strong versatility, their increased architectural complexity poses new challenges for efficiency, which remains underexplored.  
\paragraph{\textbf{Model Compression toward Parameter Efficiency}}
Despite the remarkable advances of scaling large language models, the continual growth in their size has introduced substantial redundancy and raised critical challenges for scalability. Network pruning \citep{liu2018rethinking, cheng2024survey} has emerged as an effective technique to identify and remove redundant structures in neural networks, thereby improving parameter efficiency and reducing inference cost. For instance, \cite{gromov2025the} demonstrated that many deep layers in large language models are relatively unimportant, and that comparable performance can still be maintained after removing these redundant layers. 
Within single layers, \textit{width compression} provides a complementary approach to reduce parameter count by shrinking the intermediate or hidden dimensions of model components \cite{ma2023llmpruner, xia2023sheared}.
While the uni-modal compression techniques can be transferred to Vision-Language models that take multi-modal inputs and output the language responses via language models \citep{sung2024ecoflap, Lin2024MoPECLIPSP,  10.1145/3764944.3764948}, 
it is unclear whether such methods still work in unified models. We take the prior efforts to systematically explore and exploit redundancy in multimodal models, where heterogeneous components play distinct roles. 
This perspective enables us to design compression strategies better aligned with the unified nature of multimodal understanding and generation. 
\section{Preliminaries: Unified Multimodal Models} 

Unified multimodal models are model architectures that integrate both multimodal understanding and generation within a single framework. 

\paragraph{\textbf{Understanding }} 
Let $\mathbf{x}$ denote the multimodal input, and $\mathbf{y}$ denote the corresponding output. 
The model predicts textual outputs in an auto-regressive manner:
\begin{equation}
p(\mathbf{y}_{\text{und}} \mid \mathbf{x}; \theta_{\text{und}})
= \prod_{t=1}^{T} p\!\left(y_t \mid y_{<t}, \mathbf{x}; \theta_{\text{und}}\right),
\end{equation}
where $\theta_{\text{und}}$ denotes the parameters of the understanding component, and $\mathbf{y}_{\text{und}}$ denotes the sequence of output tokens, i.e., $(y_1, y_2, \dots, y_T)$. 

\paragraph{\textbf{Generation }}
For generation tasks, the unified model leverages the understanding component to process an
instructional input $\mathbf{x}_{\text{inst}}$ (e.g., text prompt and reference images), producing conditional features 
$f_{\text{und}}(\mathbf{x}_{\text{inst}}; \theta_{\text{und}})$.
The generative component, parameterized by $\theta_{\text{gen}}$, synthesizes the output $\mathbf{y}_{\text{gen}}$, conditioned on the representation $f_{\text{und}}(\mathbf{x}_{\text{inst}}; \theta_{\text{und}})$ from the understanding component, together with an additional generative input $\mathbf{z}$ (e.g., random noise in diffusion models):
\begin{equation}
\mathbf{y}_{\text{gen}} \sim 
p\!\left(\mathbf{y}_{\text{gen}} \mid f_{\text{und}}(\mathbf{x}_{\text{inst}}; \theta_{\text{und}}), 
\mathbf{z}; \theta_{\text{gen}}\right),
\end{equation}
In unified multimodal models, the understanding component processes inputs from all modalities, while the generation component explicitly operates on non-text outputs such as image or audio synthesis. Due to their heterogeneous objectives and computational pathways, the understanding and generation components could exhibit different characteristics under compression.

\section{Methodology}

\begin{figure*}[t]
    \centering    \includegraphics[width=0.95\linewidth]{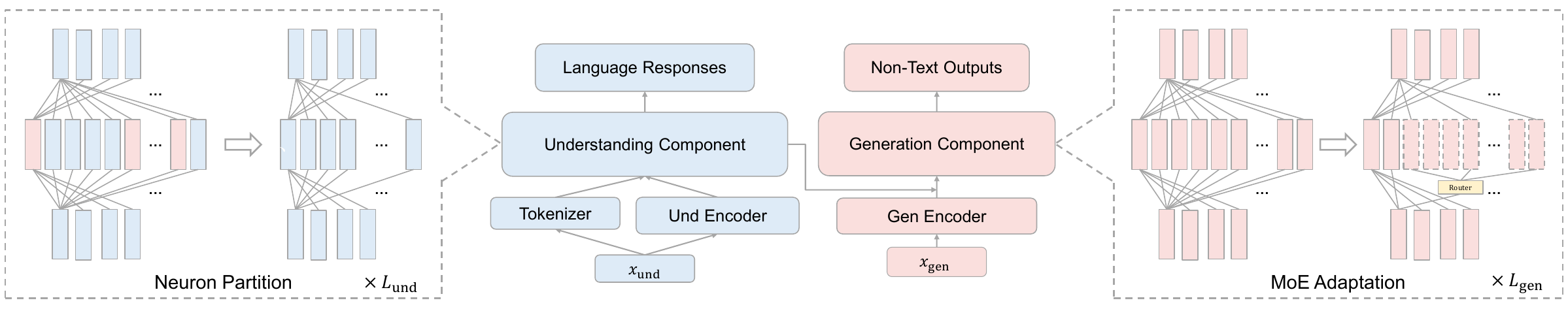}
\vspace{-5pt}
\caption{\textbf{Overview of the proposed slimness- and sparsity-oriented techniques for efficient unified multimodal models.}
We introduce two complementary strategies:
\textbf{(1) Training-free compression}, exemplified by neuron partitioning, which groups hidden neurons into subsets and prunes those less important for the target task; and
\textbf{(2) Training-aware MoE Adaptation}, which dynamically activates neurons through a Mixture-of-Experts design, where neurons are organized into shared experts (solid lines) and routed experts (dotted lines) controlled by a router. }
\label{fig:overview}
\vspace{-10pt}
\end{figure*}

In this section, we first introduce training-free pruning techniques to probe architectural redundancy, and then propose a training-aware MoE adaptation that leverages dynamic sparsity to maintain performance under sparse activation. The proposed methods are illustrated in Figure~\ref{fig:overview}.  

\subsection{Training-free Compression Strategies}

\paragraph{\textbf{Depth Pruning via Layer Dropping}}

Transformer-based large language models consist of multiple layers, and scaling their depth effectively improves performance. Nevertheless, excessive depth also introduces structral redundancy. Following \cite{gromov2025the, men2024shortgptlayerslargelanguage, he2024matterstransformersattentionneeded}, we measure the redundancy of a single layer via: 
\begin{equation}
    S_l = \text{CosineSim}(\mathbf{x}_l, \mathbf{y}_l), 
\end{equation}
where $\mathbf{x}_l$ and $\mathbf{y}_l$ correspond to the input and output of the $l$-th layer, respectively. The similarity provides a measure of redundancy, with higher values implying that the layer contributes only marginal transformation. 
The metric has been shown to perform effectively in unimodal LLMs such as Mistral \citep{jiang2023mistral7b} and LLaMA \citep{touvron2023llama2openfoundation, grattafiori2024llama3herdmodels}, and we next extend this metric to unified models. 

\paragraph{\textbf{Width Reduction via Neuron Partition}}  
In addition to depth, scaling the width, particularly within MLP layers, has become a prevalent strategy for enhancing model capability. 
In Transformer MLP layers, the input is expanded from dimension $d$ to $dm$, activating $dm$ hidden neurons before being projected back to $d$. While this expansion increases model capacity, it also introduces redundancy among neurons. To mitigate this redundancy, we propose neuron partition that separates the hidden neurons into important and less important subsets, preserving the former while pruning the latter to achieve width reduction without significantly affecting performance. 

To measure neuron importance, we take inspiration from Wanda \citep{sun2024a}, which combines weight magnitudes and activation statistics as pruning criteria, and extend this idea from unstructured weight-level pruning to a structured neuron-level metric. Given an input $x \in \mathbb{R}^{s \times d}$, in a Gate-Up-Down MLP, the hidden activations $h \in \mathbb{R}^{s \times dm}$ and output $y \in \mathbb{R}^{s \times d}$ can be written as:  
\begin{equation}
h = \big(\text{SiLU}(x W_g^\top)\big) \odot (x W_u^\top), 
\qquad 
y = h W_d^\top ,
\end{equation}
where 
$W_g, W_u \in \mathbb{R}^{m d \times d}$ are the up-projection matrix and gate-projection-matrix, 
$h \in \mathbb{R}^{n \times m d}$ is the gated activation, $W_d \in \mathbb{R}^{d \times m d}$ is the down-projection matrix. The hidden activations consist of $dm$ neurons, and the contribution of the $i$-th neuron to the final output is: 
\begin{equation}
\Delta y_i = h_i W_{d,i}^\top  ,
\end{equation}
with $W_{d,i}$ being the $i$-th column vector of $W_d$. If the $i$-th neuron is pruned, the induced output error norm can be approximated by:  
\begin{equation}
\|\Delta y\|_2 \approx \| h_i W_{d,i}^\top  \|_2 .
\end{equation}

Given all inputs from the calibration dataset $\mathcal{D}$, the expected accumulated error of each neuron is used as its importance metric: 
\begin{equation}
s_i = \mathbb{E}_{x \sim \mathcal{D}} \big[ |h_i| \cdot \|W_{d,i}^\top\|_2 \big],
\label{eq:imp}
\end{equation}
where $|h_i|$ measures the average activation magnitude of the $i$-th neuron, and $\|W_{d,i}\|_2$ quantifies its amplification effect on the output. Therefore, neurons with larger scores play more critical roles, while those with smaller scores tend to be removed. Unlike unstructured pruning that zeroes individual weights, our approach enforces structured pruning by removing entire neurons, Concretely, this corresponds to removing column $i$ from $W_d$ and row $i$ from both $W_u$ and $W_g$, thereby ensuring hardware-friendly efficiency. 

\begin{figure}[t]
  \centering
\includegraphics[width=0.9\linewidth]{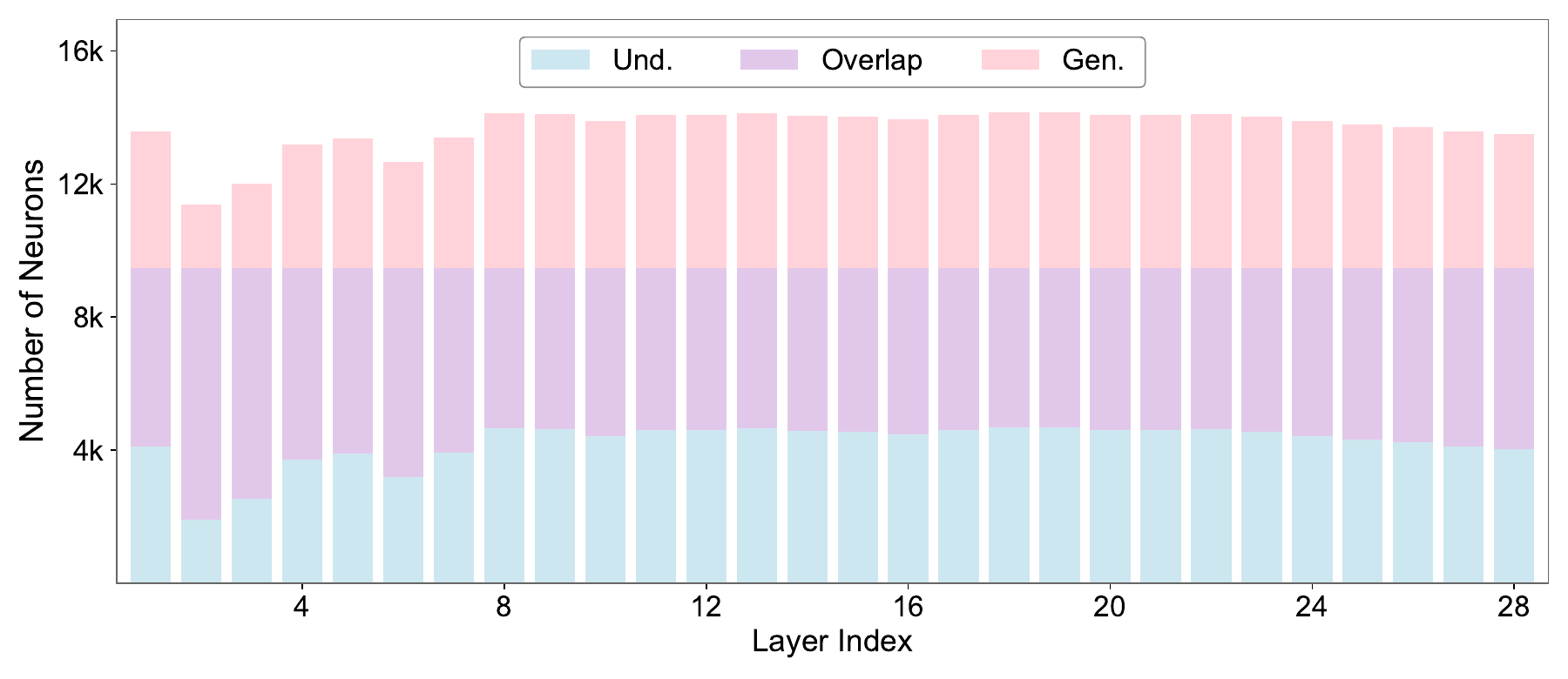} 
    \vspace{-8pt}         
\caption{\textbf{Statistical analysis of high-importance neurons}, quantifying those predominantly activated in understanding tasks, in generation tasks, and jointly across both.}
\label{fig:overlap}
\vspace{-10pt}
\end{figure}

Unified models unify diverse tasks within a single architecture, and different tasks naturally activate different subsets of neurons. 
Figure~\ref{fig:overlap} illustrates the distinct neuron partitions within the understanding component.
We first identify the top 50\% neurons according to their importance scores for understanding and for generation, respectively. Based on these two sets, we then partition neurons into three disjoint groups: those that lie only in the understanding top-50\% set, those that lie only in the generation top-50\% set, and those that are shared by both.  The relatively low overlap ratio suggests that different tasks activate distinct neuron subsets within the understanding component.
Therefore, we propose to align the calibration samples with the target tasks before applying the neuron-level importance metric, ensuring a more accurate identification of principal neurons. 

\subsection{Training-Aware MoE Adaptation}

\begin{figure}[t]
  \centering
\includegraphics[width=0.9\linewidth]{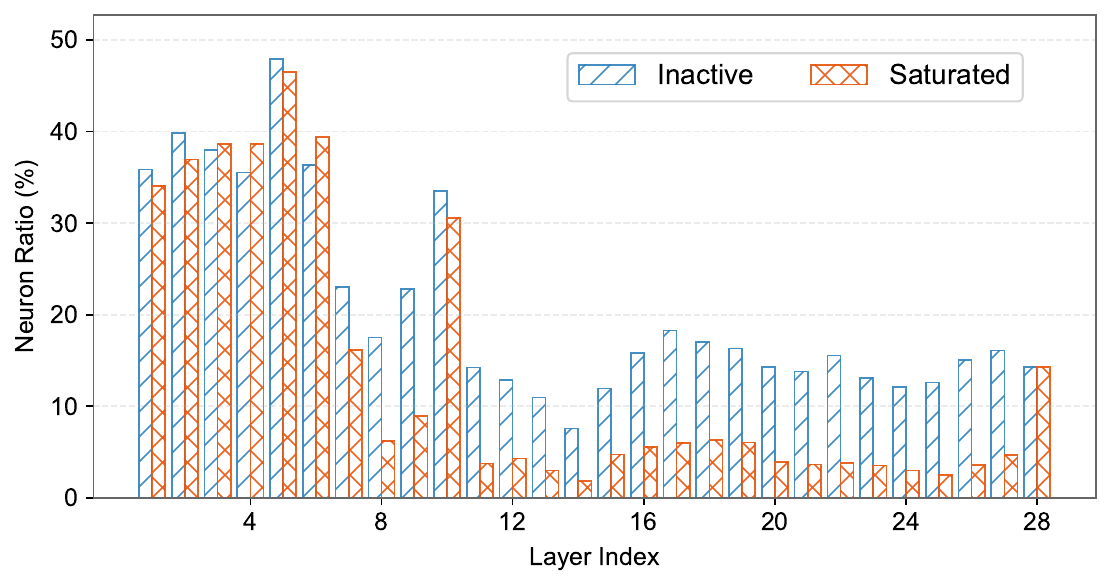} 
  \vspace{-3pt}         
\caption{\textbf{Visualization of the proportion of inactive and saturated neurons across layers. }
Active neurons are defined as those consistently ranked within the top 50\% by activation scores, while inactive neurons never enter the top 50\%. }
  \label{fig:active_inactive}
\end{figure}

Figure~\ref{fig:active_inactive} visualizes the active neurons (those consistently ranked within the top 50\% by activation scores) and inactive neurons (those never entering the top 50\%) across layers of the generation component over multiple prompts and time steps.
We observe that only a small subset of neurons remain consistently active throughout inference, while most exhibit sample-dependent activation patterns, indicating dynamic specialization across inputs. 
This observation reveals a dynamic activation phenomenon, where the subset of activated parameters depends on the input, aligning with the intuition behind the Mixture-of-Experts (MoE) architecture. To leverage this property, we incorporate an MoE mechanism into unified models through two key stages: \textit{Expert Partition} and \textit{MoE Adaptation}. 

\paragraph{\textbf{Expert Partition}} 
To separate universal and sample-specific capacity \cite{he-etal-2023-pad, dai2024deepseekmoeultimateexpertspecialization}, we partition MLP neurons into shared and routed experts using cumulative importance across calibration samples. 
For each neuron, we compute its cumulative importance score 
using Equation~\ref{eq:imp}. 
The neurons with the higher scores than the threshold are selected as shared experts $E_s$, preserving features that consistently benefit multiple samples. 
The remaining neurons, whose relateive importance is more sample-dependent, are evenly allocated to routed experts ${E_r^{(1)}, \dots, E_r^{(n)}}$ by ranked importance.
Specifically, neurons are sorted in descending order of importance and sequentially assigned to experts in alternating forward and reverse order (i.e., $E_r^{(1)}$ to $E_r^{(n)}$, then $E_r^{(n)}$ back to $E_r^{(1)}$), ensuring balanced total importance across experts.  

\paragraph{\textbf{MoE Adaptation}} 
After expert partition, we insert a router per layer to dynamically select routed experts for each input. In this case, the output of an MoE layer is formulated as follows: 
\begin{equation}
    \text{MoE}(x) = f_{\mathcal{S}}(x) + \sum_{j \in \text{Top-}k(\mathcal{G})} \mathcal{G}_j \cdot f_{\mathcal{R}_j}(x), 
\label{eq:moe}
\end{equation}
where $\mathcal{G}$ denotes the gating function, and $f_{\mathcal{S}}$ and $f_{\mathcal{R}}$ represent the transformations of the shared and routed experts, respectively. The original MLP layer can be regarded as a special case of Equation~\ref{eq:moe}, where all experts are selected and their gating scores are uniformly set to 1. To enable a smooth transition from dense to sparse activation, we relax the constraint that the gating scores sum to 1 and reparameterize them as $(1 + \text{Router}(x))$, where the router network’s gate is initialized to zero. 
To adapt the model to the converted sparse activation pattern, we perform Expert-Frozen Tuning as a cold-start phase of MoE Adaptation, where experts remain fixed while the router and other parameters are trainable.
This stage allows the model to leverage the pretrained knowledge of existing experts while establishing a preliminary routing policy. Afterward, we release the freezing constraint to fully enable MoE Adaptation.

\section{Experiments}

\subsection{Experimental Setup}
\label{app:setting}

\paragraph{\textbf{Models}}
We focus on several mainstream open-source unified models, including BAGEL \citep{deng2025emergingpropertiesunifiedmultimodal}, Ming-Omni \citep{ai2025mingomniunifiedmultimodalmodel}, and Qwen-Image \citep{wu2025qwenimagetechnicalreport}.
All three adopt Qwen-Instruct \citep{qwen2} as the backbone for multimodal understanding. 
For the understanding component, BAGEL and Qwen-Image rely on a VLM derived from Qwen-Instruct~\citep{qwen2}, whereas Ming-Omni employs an MoE-based backbone  \cite{ling}. 
The generation components further diverge across models: BAGEL employs a Mixture-of-Transformers (MoT) \citep{liang2025mixtureoftransformers} design and reuses the Qwen-Instruct backbone for generation; Qwen-Image incorporates an MMDiT-based generator \citep{DBLP:conf/icml/EsserKBEMSLLSBP24} and Ming-Omni adopts a multi-scale DiT block architecture. Table~\ref{tab:models} presents a detailed comparison of these models. 

\begin{table}[h]
\centering
\caption{{Summary of evaluated unified models.}}
\resizebox{\linewidth}{!}{
\begin{tabular}{l|c|c|c|c}
\toprule
\textbf{Model} & ~\textbf{Und. Component}~ & ~\textbf{Und. Param.}~ & ~\textbf{Gen. Component}~ & ~\textbf{Gen. Param.}~ \\
\midrule
Qwen-Image & VLM & 7.62B & MMDiT & 20.42B \\
Ming-Omni & MoE-VLM & 17.12B & MMDiT
 & 2.51B \\
BAGEL & VLM & 7.62B & LLM & 7.62B \\
\bottomrule
\end{tabular}
}
\label{tab:models}
\end{table}

We systematically vary the number of experts per MoE layer ({16, 32, 64}). Following the design choices in \cite{dai2024deepseekmoeultimateexpertspecialization, deepseekai2025deepseekv3technicalreport}, we include shared experts that constitute one-sixteenth of the total number of experts. The overall activation ratio is set to $50\%$ per layer. We exclude the first and last layers from MoE conversion, as they are essential for preserving input encoding and output generation quality \cite{dai2024deepseekmoeultimateexpertspecialization}. 

\paragraph{\textbf{Datasets}}
For the calibration datasets used in training-free compression, 
we use a small number of examples drawn directly from the target task, which is sufficient for training-free compression \cite{Frantar2022GPTQAP, Lin2023AWQAW,men2024shortgptlayerslargelanguage, he2024matterstransformersattentionneeded}. 
Note that these samples are used solely to compute activation scores, so no ground-truth annotations are required and no label leakage occurs.
For MoE adaptation, we additionally incorporate high-quality image–text pairs, complemented by a small amount of synthetic data generated by existing text-to-image models.

\subsection{Redundancy of Und. Component}

\begin{figure}[h]
  \centering
\includegraphics[width=0.9\linewidth]{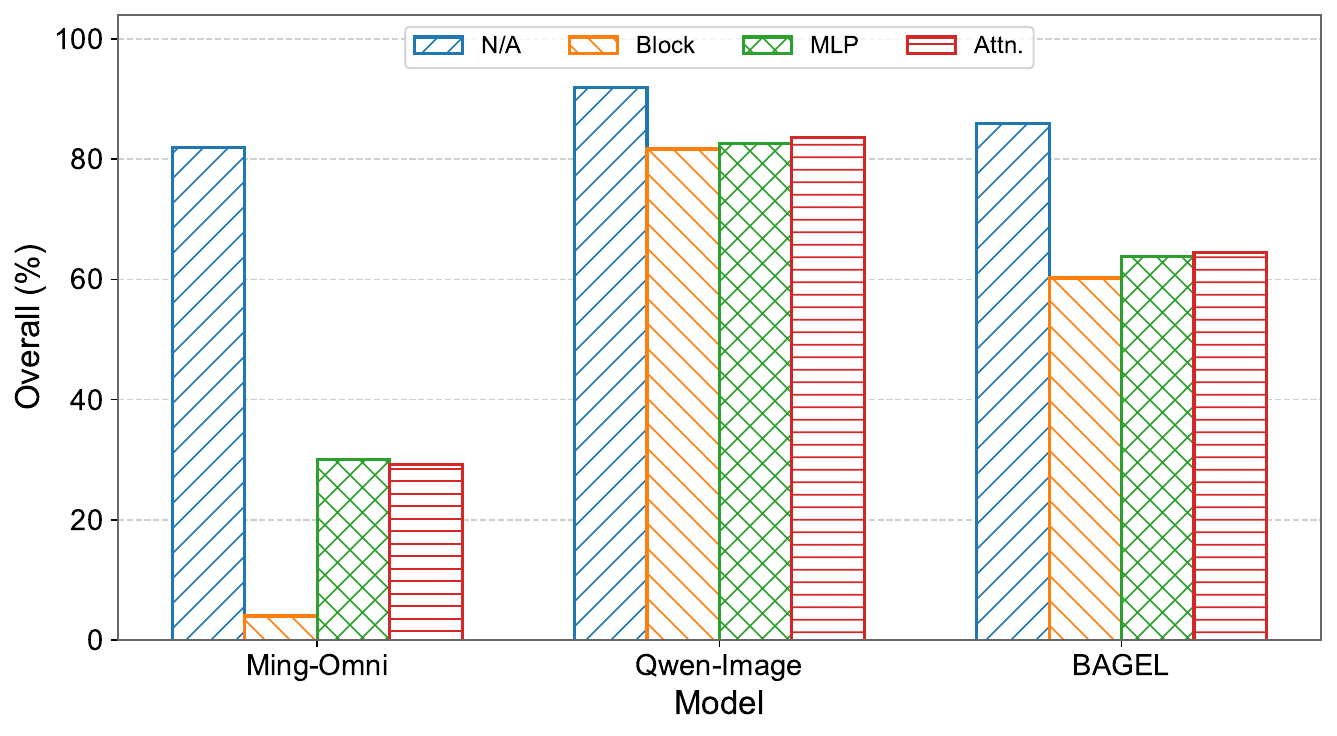} 
  \vspace{-10pt}
  \caption{Comparison of the overall performance of depth reduction on the GenEval, where the compression ratio is set as 50\%. }
  \vspace{-5pt}
\label{fig:depth_reduction}
\end{figure}

\paragraph{\textbf{Depth Reduction works in Generation Tasks but Fails in Understanding}} 
Since the understanding component is not directly connected to the generation output, we first analyze its indirect influence on generation performance. Specifically, we remove transformer blocks, MLP layers, and attention layers, respectively. 
As shown in Figure~\ref{fig:depth_reduction}, removing $50\%$ of layers in the understanding component proves effective for BAGEL and Qwen-Image, but is less effective for Ming-Omni. We attribute this difference to architectural design: Ming-Omni’s generation component is relatively smaller and thus depends more heavily on precise features encoded by the understanding component.

However, depth pruning substantially deteriorates the model’s understanding capability. 
For instance, removing half of the MLP layers causes performance on MME~\citep{Fu2023MMEAC} to drop from 1684.8 to 304.5 in perception and from 696.7 to 127.1 in cognition. These results suggest that depth reduction fails to preserve the performance of unified multimodal models on understanding tasks. Language-response-oriented understanding tasks rely on autoregressive decoding, which is inherently an error accumulation process, where the deviation of previous timesteps can propagate through subsequent decoding steps and ultimately cause the model to collapse within just a few steps. 

\begin{table}[h]
\vspace{-10pt}
\centering
\caption{Performance of neuron partitioning on understanding tasks at compression ratios of 25\% and 50\%.} 
\vspace{-5pt}
\resizebox{\linewidth}{!}{
\begin{tabular}{l|c|ccccc}
\toprule
\textbf{Model} & \textbf{Sparsity} & \textbf{MME-P} & \textbf{MME-C} & \textbf{MMMU} & \textbf{MMBench} & \textbf{MMVP} \\
\midrule
\multirow{3}{*}{Ming-Omni}
  & --   & 1584.3 & 670.4 & 66.7 & 86.7 & 54.6 \\
  & 25\% & 1578.5 & 560.4 & 56.7 & 81.2 & 51.3 \\
  & 50\% & 1269.0 & 317.9 & 51.7 & 81.0 & 46.0 \\
\midrule
\multirow{3}{*}{BAGEL}
  & --   & 1684.8 & 696.7 & 65.0 & 88.1 & 69.6 \\
  & 25\% & 1558.1 & 681.7 & 60.1 & 85.7 & 68.7 \\
  & 50\% & 1392.6  & 528.9 & 56.7 & 79.2 & 56.0 \\
\bottomrule
\end{tabular}}
\vspace{-15pt}
\label{tab:width_und}
\end{table}

\begin{table*}[t]
\centering
\caption{\textbf{Performance on GenEval when applying training-free Neuron Partition to the understanding component}. 
Since only the \textbf{understanding component} is compressed, the reported parameter counts correspond to this part rather than the full model size. }
\resizebox{\linewidth}{!}{
\begin{tabular}{l|c|c|cccccc|c}
\toprule
\textbf{Model} 
& ~\textbf{Sparsity}~ & ~\textbf{Params.}~ & ~\textbf{Single Obj.}~ & ~\textbf{Two Obj.}~ & ~\textbf{Counting}~ & ~\textbf{Colors}~ & ~\textbf{Position} & ~\textbf{Color Attri.}~ & ~\textbf{Overall}$\uparrow$ \\
\toprule
\multirow{3}{*}{{BAGEL}}  
& $0\%$ & 7.62B & 0.99 & 0.94 & 0.81 & 0.95 & 0.72 & 0.77 & \underline{0.86} \\
& $25\%$ & 6.19B & 0.98 & 0.91 & 0.81 & 0.94 & 0.70 & 0.70 & \underline{0.84} \\
& $50\%$ & 4.76B & 0.94 &	0.63	& 0.62	& 0.77	& 0.47	& 0.34	& \underline{0.63} \\
\midrule
\multirow{3}{*}{{Qwen-Image}}          
& $0\%$ & 7.62B
& 0.99 & 0.98 & 0.91 & 0.94 & 0.80 & 0.89 & \underline{0.92}  \\
&  $50 \%$ & 4.76B & 0.99 & 0.94 & 0.94 & 0.93 & 0.76 & 0.87 & \underline{0.90}
  \\
& $70 \%$ & 3.62B & 0.97 & 0.88 & 0.85 & 0.91 & 0.60 & 0.71 & \underline{0.82}
 \\

\midrule
\multirow{3}{*}{{Ming-Omni}}        
& $0\%$ & 17.12B
& 0.97 & 0.95 & 0.67 & 0.92 & 0.71 & 0.71 & \underline{0.82}  \\
& $50 \%$ & 8.55B & 0.97 & 0.92 & 0.66 & 0.89 & 0.61 & 0.70 & \underline{0.79}  \\
&  $70 \% $ & 5.61B & 0.96 & 0.81 & 0.58 & 0.86 & 0.49 & 0.56 & \underline{0.71}  \\
\bottomrule
\end{tabular}}

\label{tab:width_reduction_GenEval}
\end{table*}

\begin{figure*}[htbp]
\centering
\setlength{\tabcolsep}{3pt} 
\renewcommand{\arraystretch}{0.9} 

\resizebox{\linewidth}{!}{
\begin{tabular}{c|c|c|c|c|c}
\toprule
\textbf{Baseline} 
& \textbf{Comp. w/Gen. } 
& \textbf{Comp. w/Und. } 
&
\textbf{Baseline} 
& \textbf{Comp. w/Gen. } 
& \textbf{Comp. w/Und. } 
\\
\midrule
\includegraphics[width=0.15\textwidth]{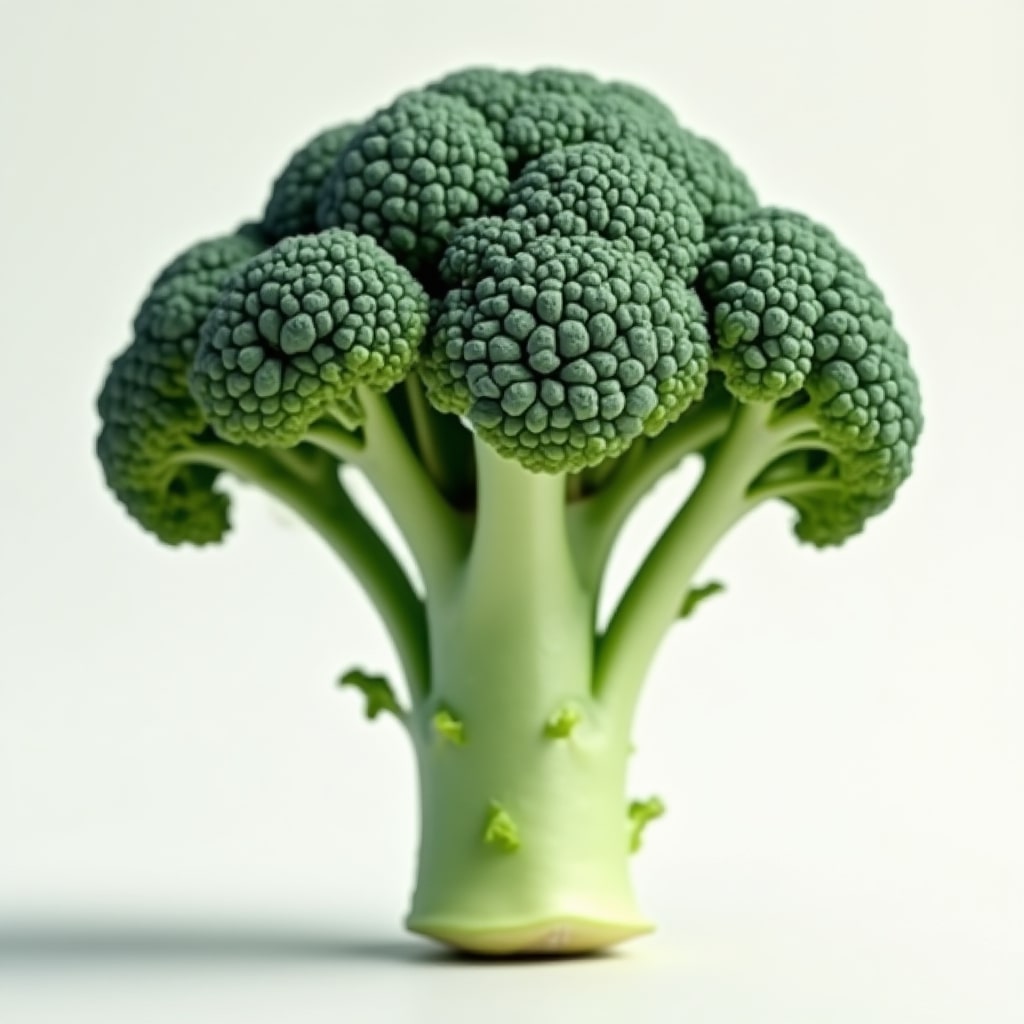} &
\includegraphics[width=0.15\textwidth]{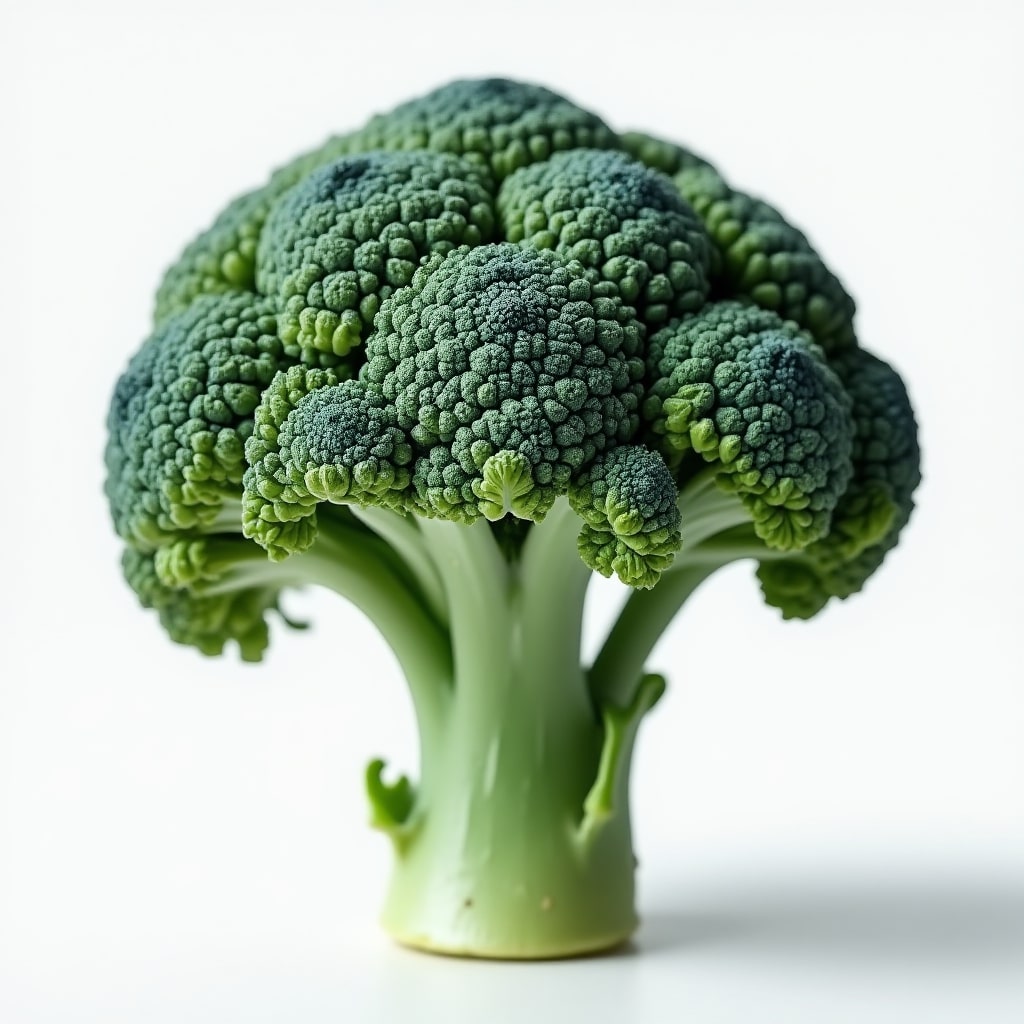} &
\includegraphics[width=0.15\textwidth]{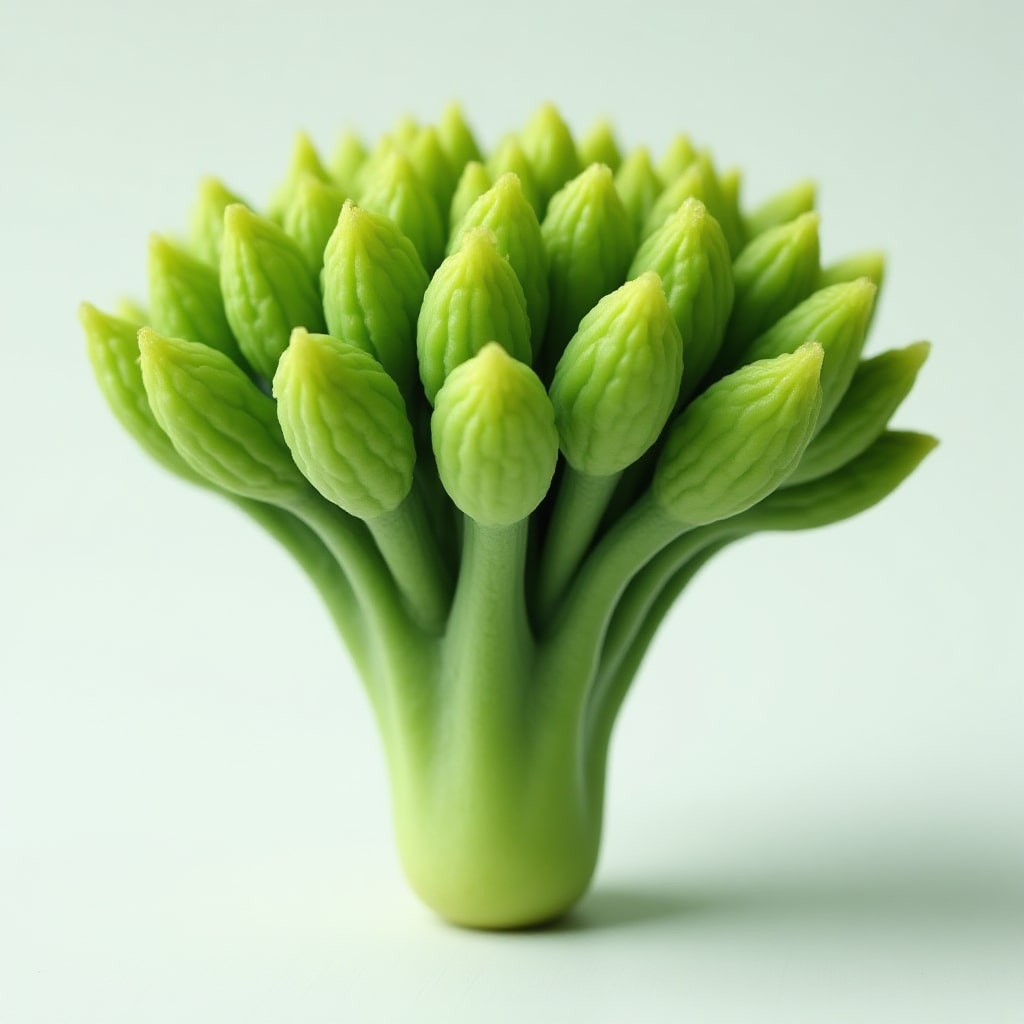} &
\includegraphics[width=0.15\textwidth]{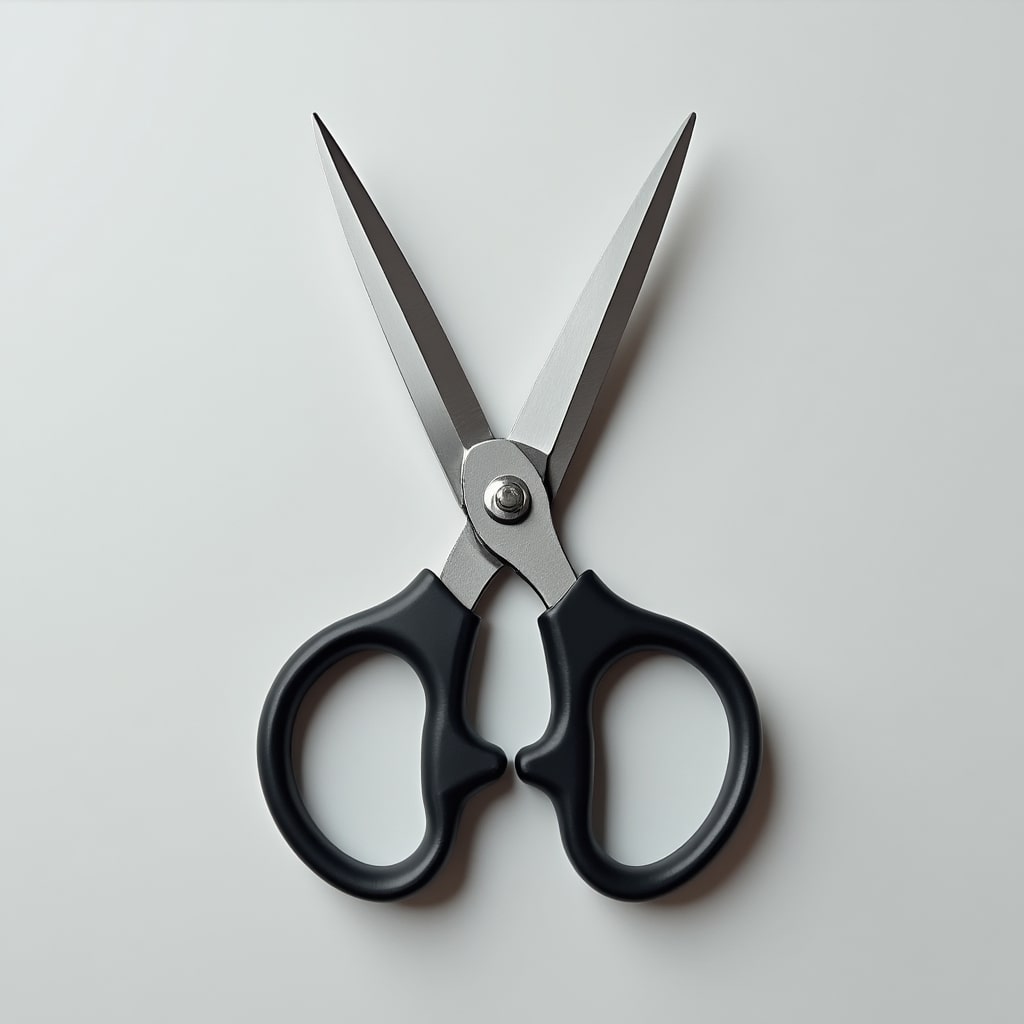} &
\includegraphics[width=0.15\textwidth]{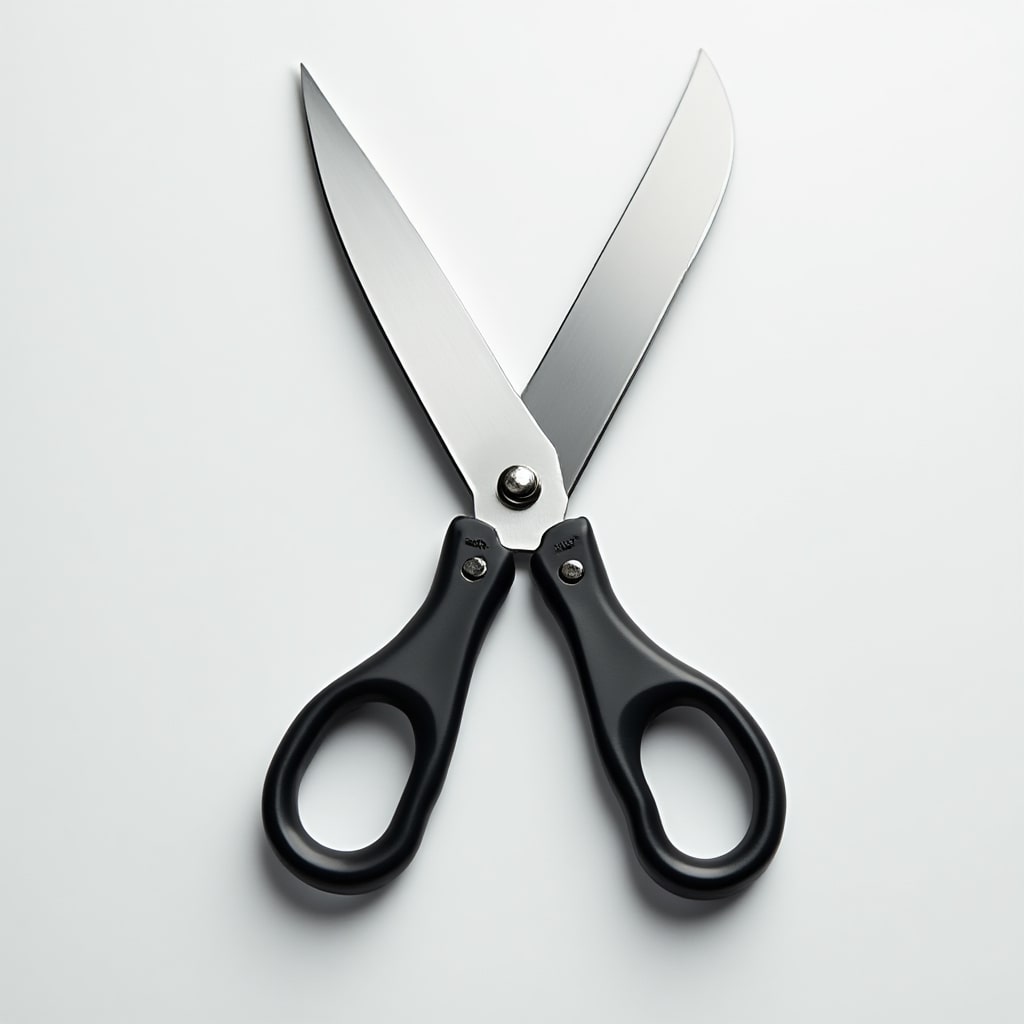} &
\includegraphics[width=0.15\textwidth]{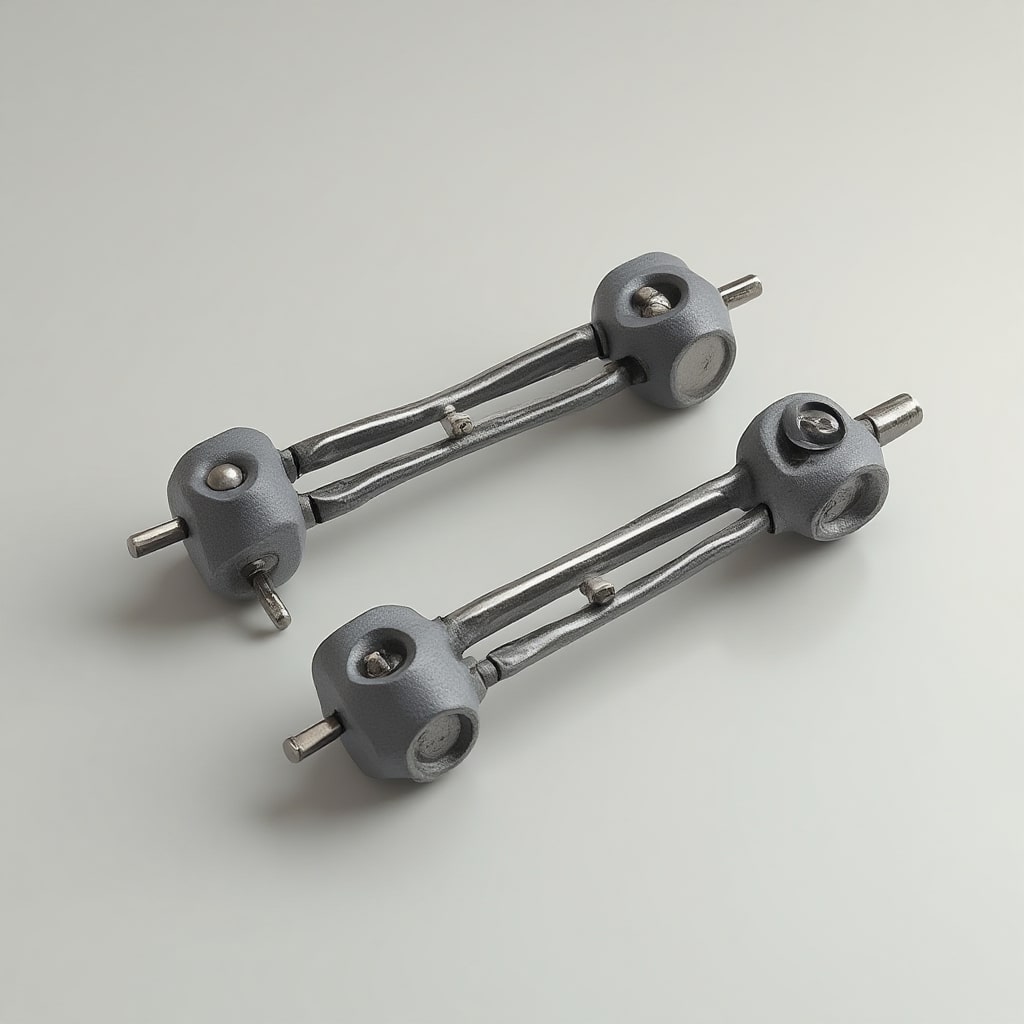} 
\\
\midrule
\multicolumn{3}{c|}{(a) A realistic broccoli sits upright on a plain surface.} & \multicolumn{3}{c}{(b) A pair of scissors lies on a flat surface.}  
\\
\midrule
\includegraphics[width=0.15\textwidth]{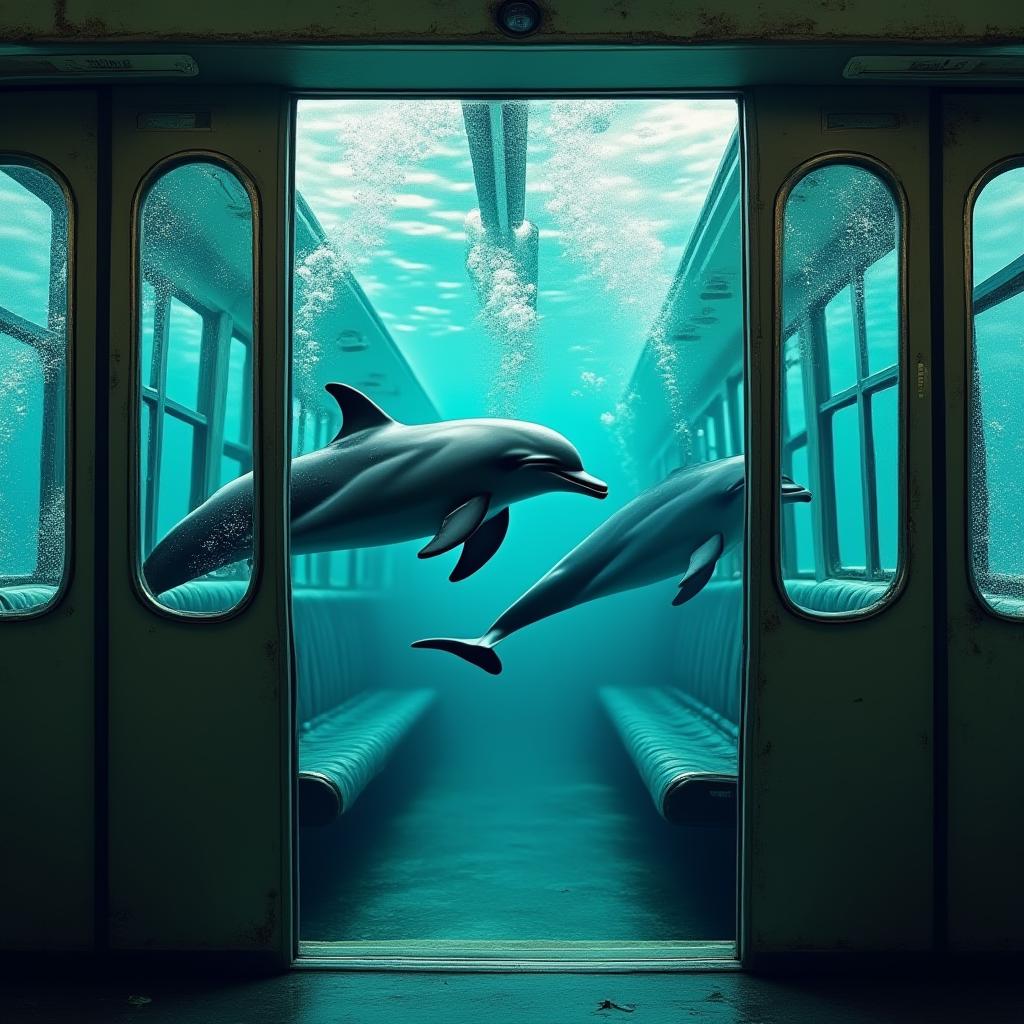} & 
\includegraphics[width=0.15\textwidth]{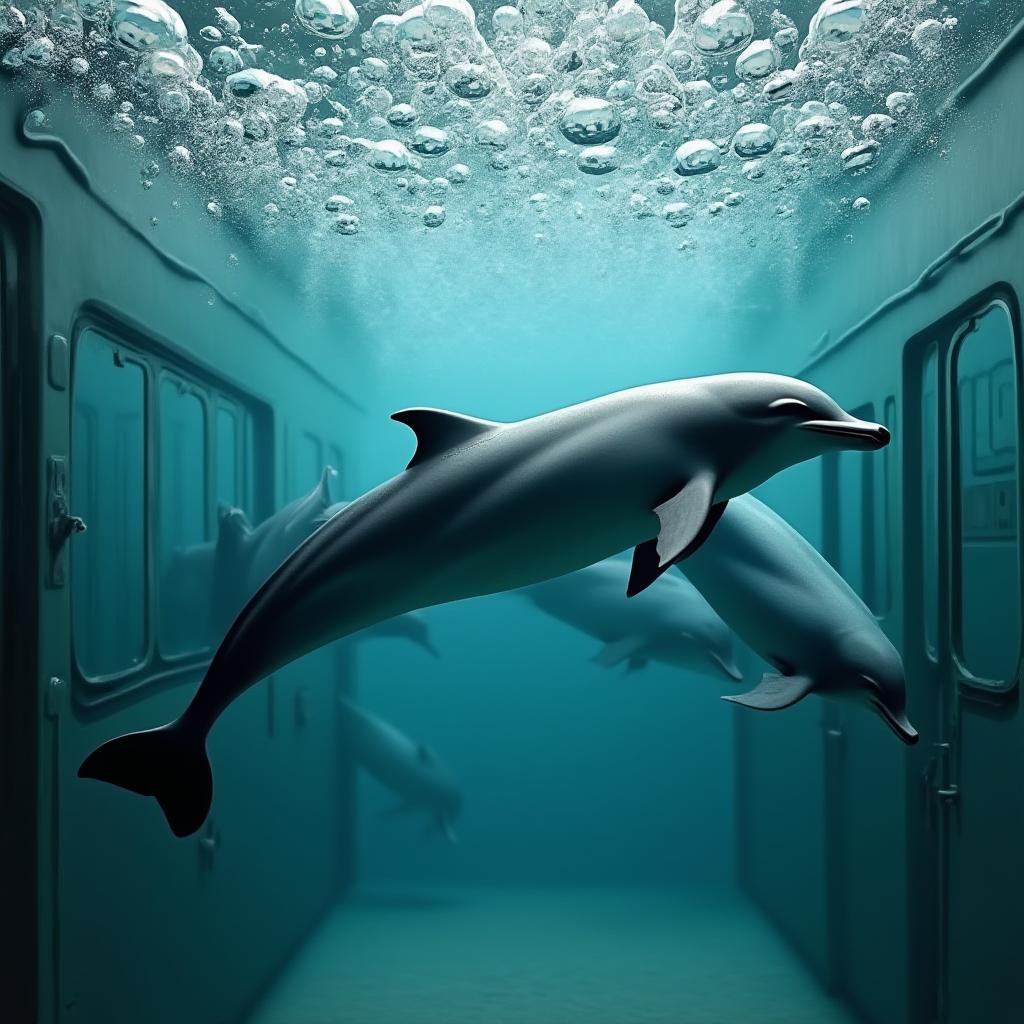} & 
\includegraphics[width=0.15\textwidth]{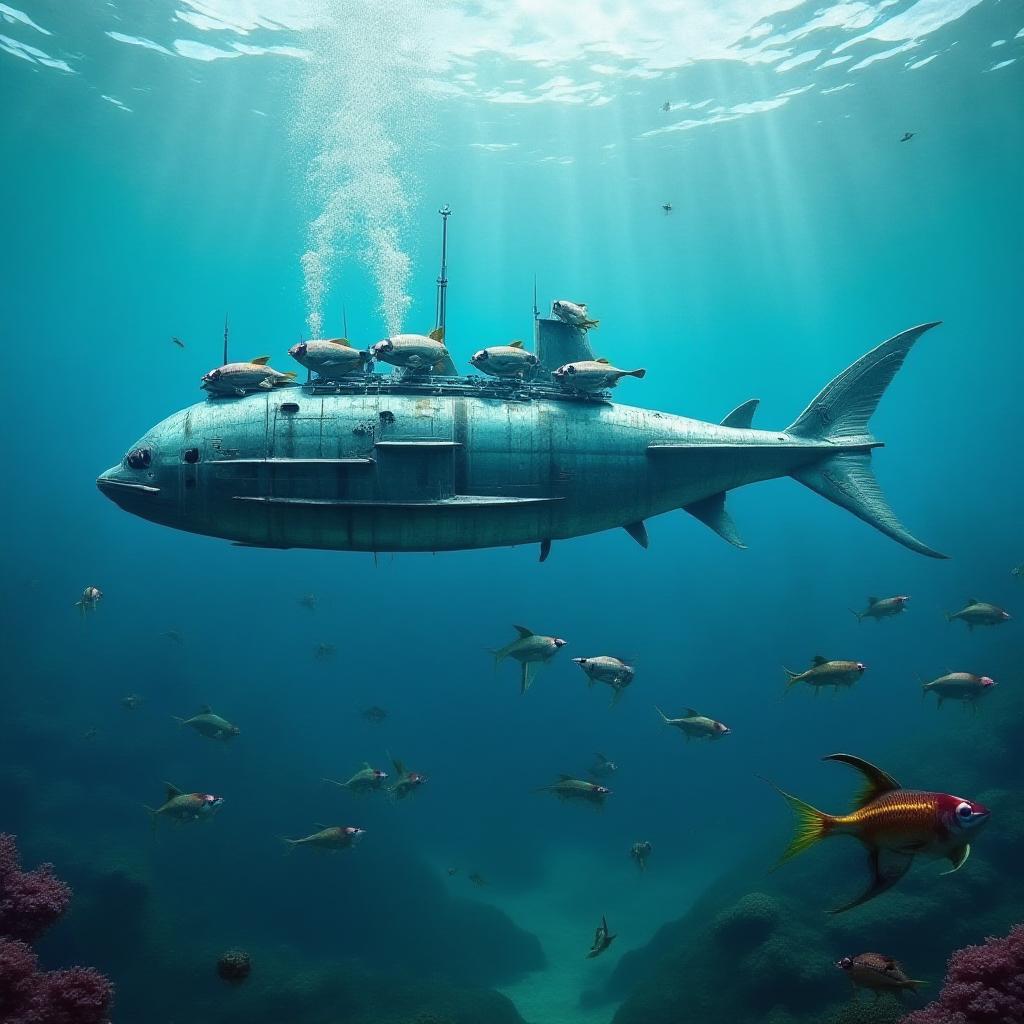} & 
\includegraphics[width=0.15\textwidth]{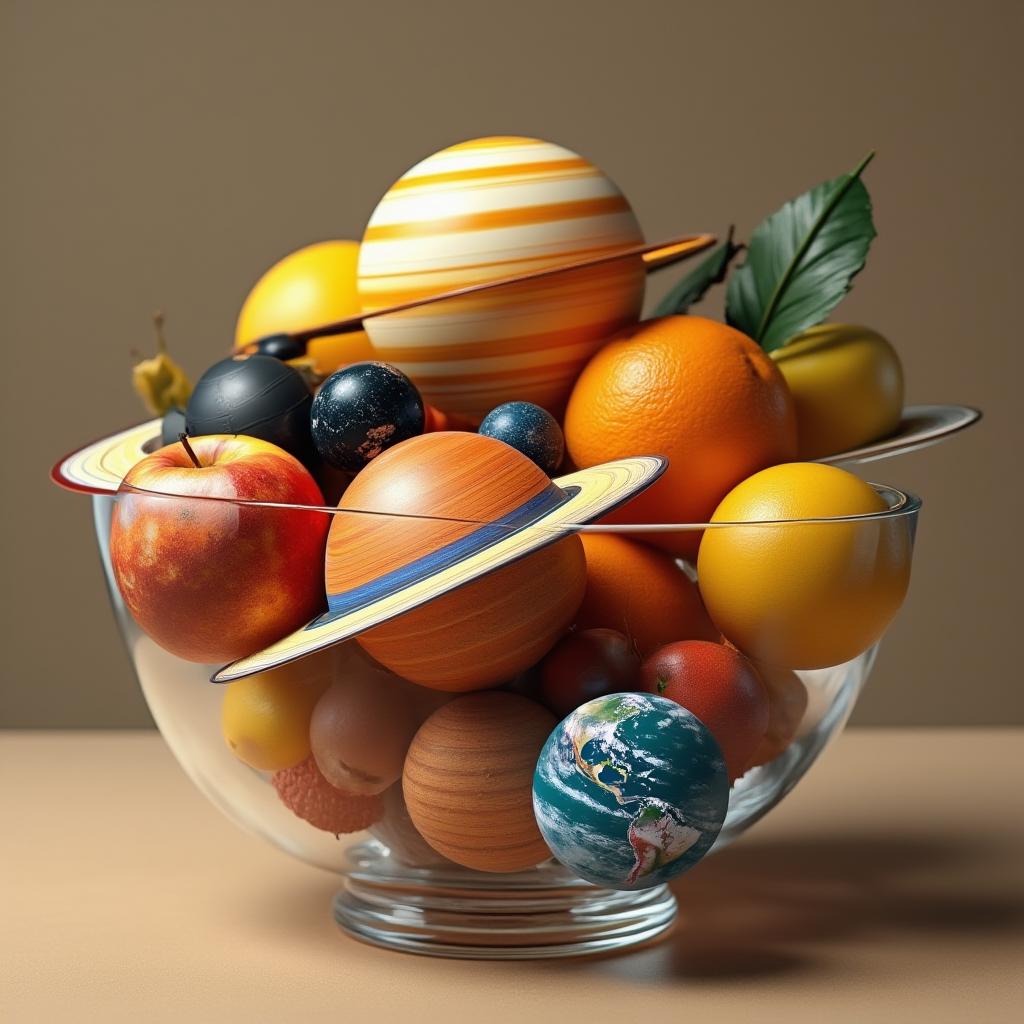} & 
\includegraphics[width=0.15\textwidth]{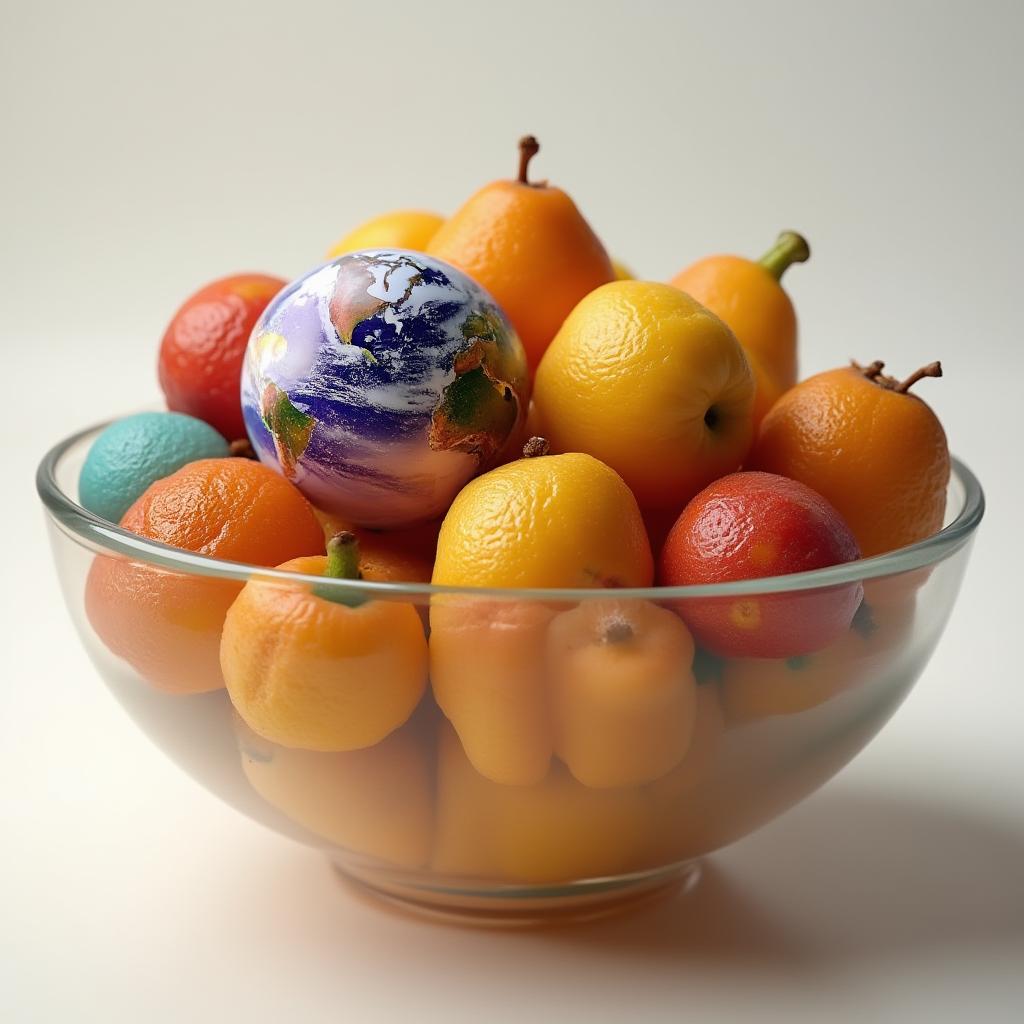} & 
\includegraphics[width=0.15\textwidth]{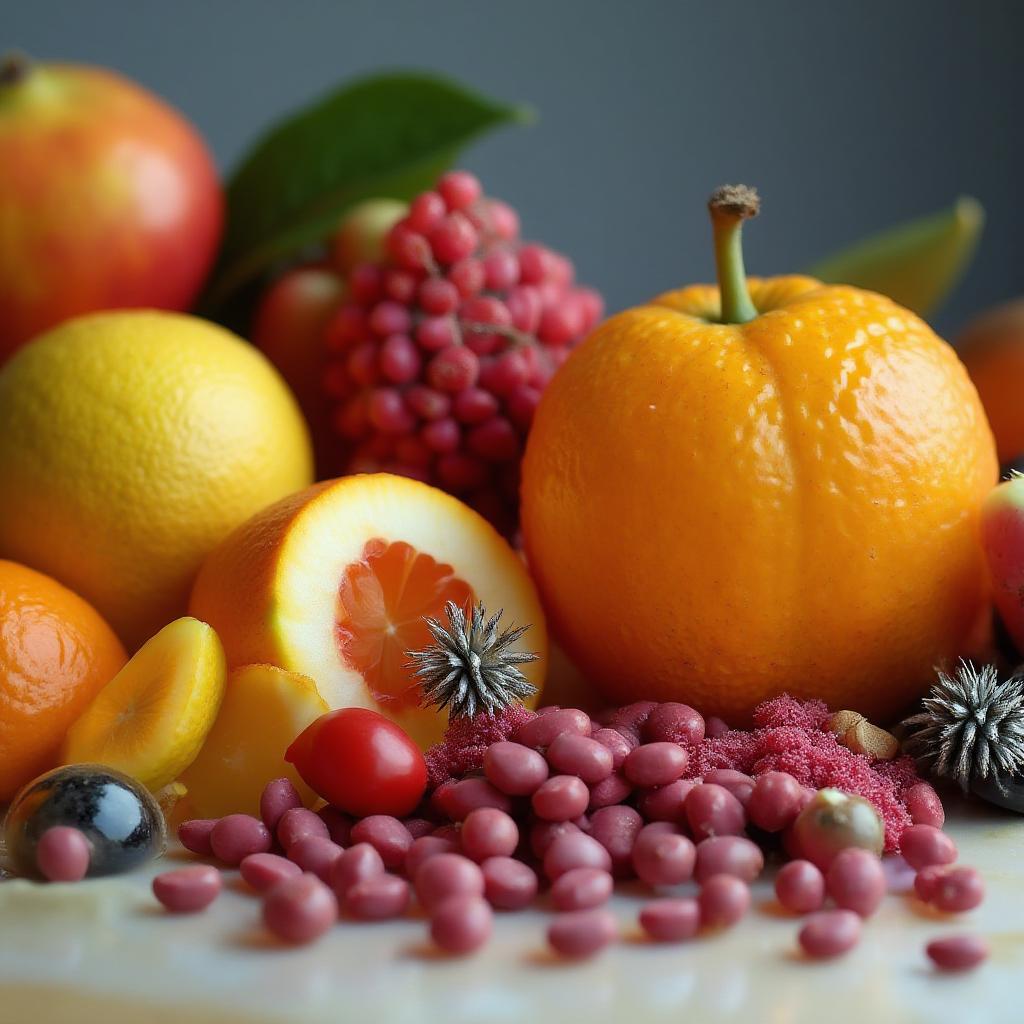} 
\\
\midrule
\multicolumn{3}{c|}{(c)  dolphins swim through abandoned subway cars. } & 
\multicolumn{3}{c}{(d)  a fruit bowl consisting of fruits and miniature planets. } 
\\
\bottomrule
\end{tabular}}

\caption{\textbf{Impact of calibration data selection on neuron partition within the understanding component for generation tasks.}
Each triplet presents outputs from the \textbf{unmodified model (left)}, the model after neuron partition using \textbf{image-generation calibration (middle)}, and using \textbf{understanding calibration (right)}.} 
\label{fig:calibration_data}
\vspace{-7pt}
\end{figure*}

\paragraph{\textbf{Neuron Partition on Understanding Components: Effective in Both Understanding and Generation}} We next evaluate the effectiveness of neuron partition on understanding components. Specifically, we compress the MLP layers to the target ratios using a small set of calibration samples. As shown in Table~\ref{tab:width_reduction_GenEval}, Ming-Omni and Qwen-Image largely maintain their performance even under aggressive compression ratios (i.e., $50\%$ and $75\%$), whereas BAGEL exhibits a greater loss in capability, likely due to its mixture-of-transformers architecture \citep{liang2025mixtureoftransformers}, in which components interact more frequently through cross-attention at every layer. 

\begin{figure*}[htbp]
\centering
\setlength{\tabcolsep}{3pt} 
\renewcommand{\arraystretch}{0.9} 

\begin{tabular}{c|c}
\toprule
\includegraphics[width=0.48\linewidth]{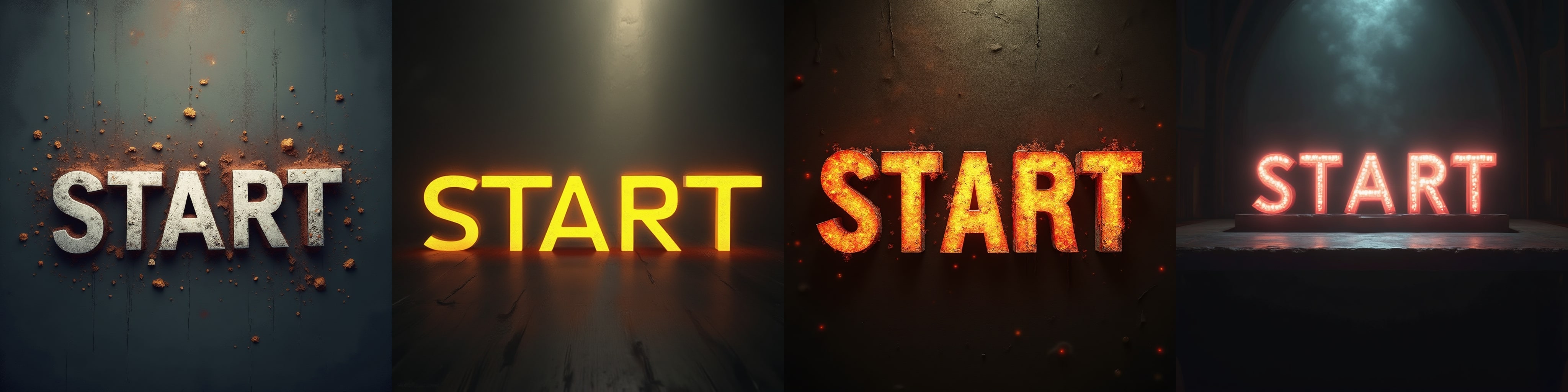} & 
\includegraphics[width=0.48\linewidth]{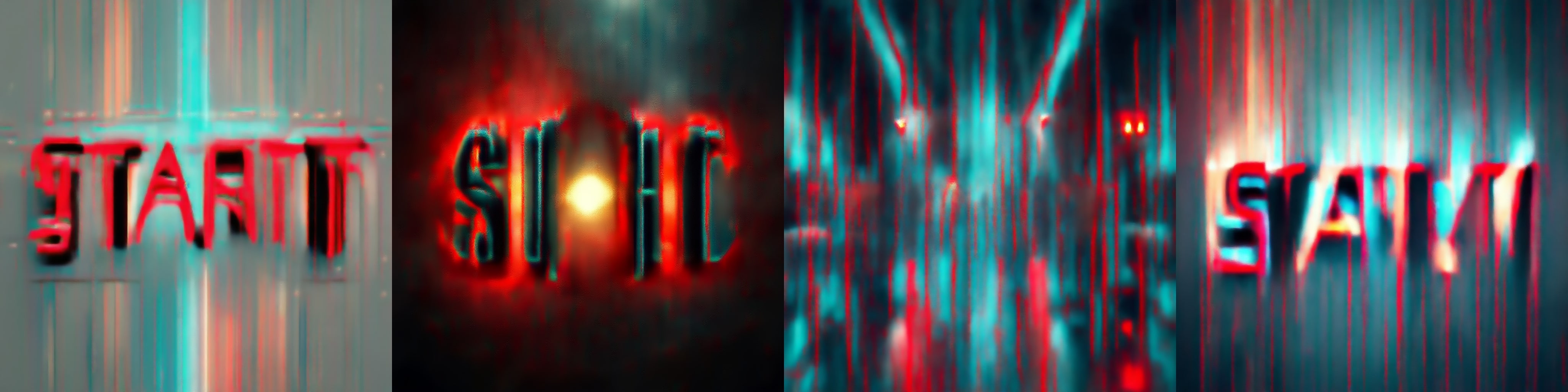} \\ 
\midrule
Baseline (uncompressed) & 
Compressed (50\% width reduction ) 
\\
\bottomrule
\end{tabular}

\caption{
\textbf{Qualitative comparison between the baseline (uncompressed) model and the model with 50\% width reduction in the generation component.}
The \textbf{baseline model (left)} is evaluated without compression, while the \textbf{compressed model (right)} reduces the generator width by 50\%.
Results are shown for the prompt “The word \texttt{START}.” Compression leads to noticeable degradation in fine details and semantic consistency.
}
\label{fig:gen_dilemma}
\vspace{-12pt}
\end{figure*}

Notably, as shown in Table~\ref{tab:width_und}, neuron partition consistently outperforms layer dropping across understanding tasks. Although certain layers exhibit substantial redundancy, aggressively removing them would also eliminate the small subset of weights that are critical to task performance~\cite{yu2025the}. In contrast, neuron partition selectively preserves important neurons within each layer that are most relevant to the target task, thereby achieving more fine-grained and reliable compression. 
Nevertheless, since the understanding component directly governs textual outputs, its compression ratio should remain more conservative in understanding tasks than in generation tasks. 

\paragraph{\textbf{Calibration Data Affects the Activated Parameters}} 
Neuron partition leverages calibration samples to estimate neuron importance and prunes those that are less critical based on their activation behavior. Figure~\ref{fig:overlap} shows that different neurons are activated for different tasks, suggesting that the choice of calibration data can lead to varying retained weights and thereby affect downstream performance. 
To assess how different calibration datasets affect parameter retention and downstream performance, we conduct an ablation study using samples from understanding tasks (MME) and generation tasks (GenEval). 

We find the alignment between calibration data and target tasks contributes to the performance. 
For instance, 
Using calibration samples from the understanding and generation tasks yields MMBench scores of 79.2 and 74.8, respectively. The generation results in Figure~\ref{fig:calibration_data} 
further highlight this trend. When calibrated with image generation samples, the outputs remain faithful to the prompts, producing broccoli, scissors, dolphins, and fruit bowls with correct structures. In contrast, calibration with understanding samples introduces distortions and mismatches. 
This demonstrates that task-aligned calibration data yields better performance, while mismatched data degrades generation quality. The effect is particularly critical for unified models, where both input and output types vary in different combination of modalities. 

\subsection{Dilemma of Gen. Component Compression}  

We next investigate how compression influences generation quality by applying neuron partition 
to the generation components. 
While neuron partition demonstrates strong performance in compressing the understanding components, applying similar compression to the generation components presents a clear dilemma. As illustrated in Figure~\ref{fig:gen_dilemma}, aggressive compression, such as a $50\%$ width reduction, substantially degrades the fidelity and coherence of generated outputs. Compressed models often produce distorted structures and unrealistic textures, deviating from the intended semantics. 
This highlights the contrasting compressibility between understanding and generation components: whereas understanding tasks remain robust under compression, generation components are highly sensitive, limiting the extent of feasible compression.

 \begin{table*}[thbp]
\centering

\caption{\textbf{Performance comparison across stages of MoE Adaptation}: Expert Partition (no training), Expert-Frozen Tuning, and full MoE Adaptation. 
We evaluate Expert Partition (without training), Expert-Frozen Tuning, and full MoE Adaptation across two configurations of adapted components (\textit{Gen.} and \textit{Und. \& Gen.}).
For reference, results from the dense model pruned with neuron partition and subsequent fine-tuning under an equivalent budget of activated parameters are also reported. Activated Params. denote the number of activated parameters in the understanding and generation components, shown in the format ``{Und. Param. } \& {Gen. Param.}''.  }
\resizebox{0.98\linewidth}{!}{
\begin{tabular}{l|c|c|cccccc|c}
\toprule
\textbf{Method} & 
~\textbf{Adapt. Comp.}~ & 
~\textbf{Activated Params.}~ & ~\textbf{Single Obj.}~ & ~\textbf{Two Obj.}~ & ~\textbf{Counting}~ & ~\textbf{Colors}~ & ~\textbf{Position}~ & ~\textbf{Color Attri.}~ & ~\textbf{Overall}$\uparrow$ \\

\midrule
Baseline & N/A & 7.62B + 7.62B & 0.99 & 0.94 & 0.81 & 0.95 & 0.72 & 0.77 & \underline{0.86} \\
\midrule
 Expert Partition & \multirow{4}{*}{\textit{Gen.}}  & \multirow{4}{*}{7.42B + 4.96B} & 0.90 & 0.70 & 0.49 & 0.74 & 0.53 & 0.34 & \underline{0.62} \\
 Dense Finetuning &   &  & 0.97 & 0.88 & 0.75 & 0.91 & 0.67 & 0.71 & \underline{0.82} \\

 Expert-frozen Tuning~~ & & & 0.99 & 0.94 & 0.62 & 0.93 & 0.69 & 0.54 & \underline{0.78} \\
MoE Adaptation &  &  
& \bf 0.99 & \bf 0.95 & \bf 0.85 & \bf 0.95 & \bf 0.75 & \bf 0.79 & \bf \underline{0.88}
 \\
 \midrule
Expert Partition & 
\multirow{4}{*}{\textit{Und. $\&$ Gen.}}  
& \multirow{4}{*}{4.96B + 4.96B} & 0.69 & 0.18 & 0.23 & 0.45 & 0.10 & 0.05 & \underline{0.28}
\\

 Dense Finetuning &   &  & 0.97 & 0.89 & 0.76 & 0.91 & 0.70	& 0.64 & \underline{0.81} \\

Expert-frozen Tuning & & & 0.94 & 0.63 & 0.62 & 0.77 & 0.47 & 0.34	& \underline{0.63} \\

MoE Adaptation &  &  & \bf 0.99 & \bf 0.96 & \bf 0.78 & \bf 0.95 & \bf 0.70 & \bf 0.72 & \bf \underline{0.85} \\

\bottomrule
\end{tabular}}
\label{tab:router-tuning}
\end{table*}

\subsection{MoE Adaptation for Gen. Component Sparsity} 
Static compression inherently conflicts with the dynamic activation patterns required across tasks and samples, leading to notable degradation in generation components. To mitigate this issue, we apply MoE Adaptation, enabling dynamic activation to restore representational capacity and overall performance. 

\begin{figure}[ht]
  \centering
\includegraphics[width=0.9\linewidth]{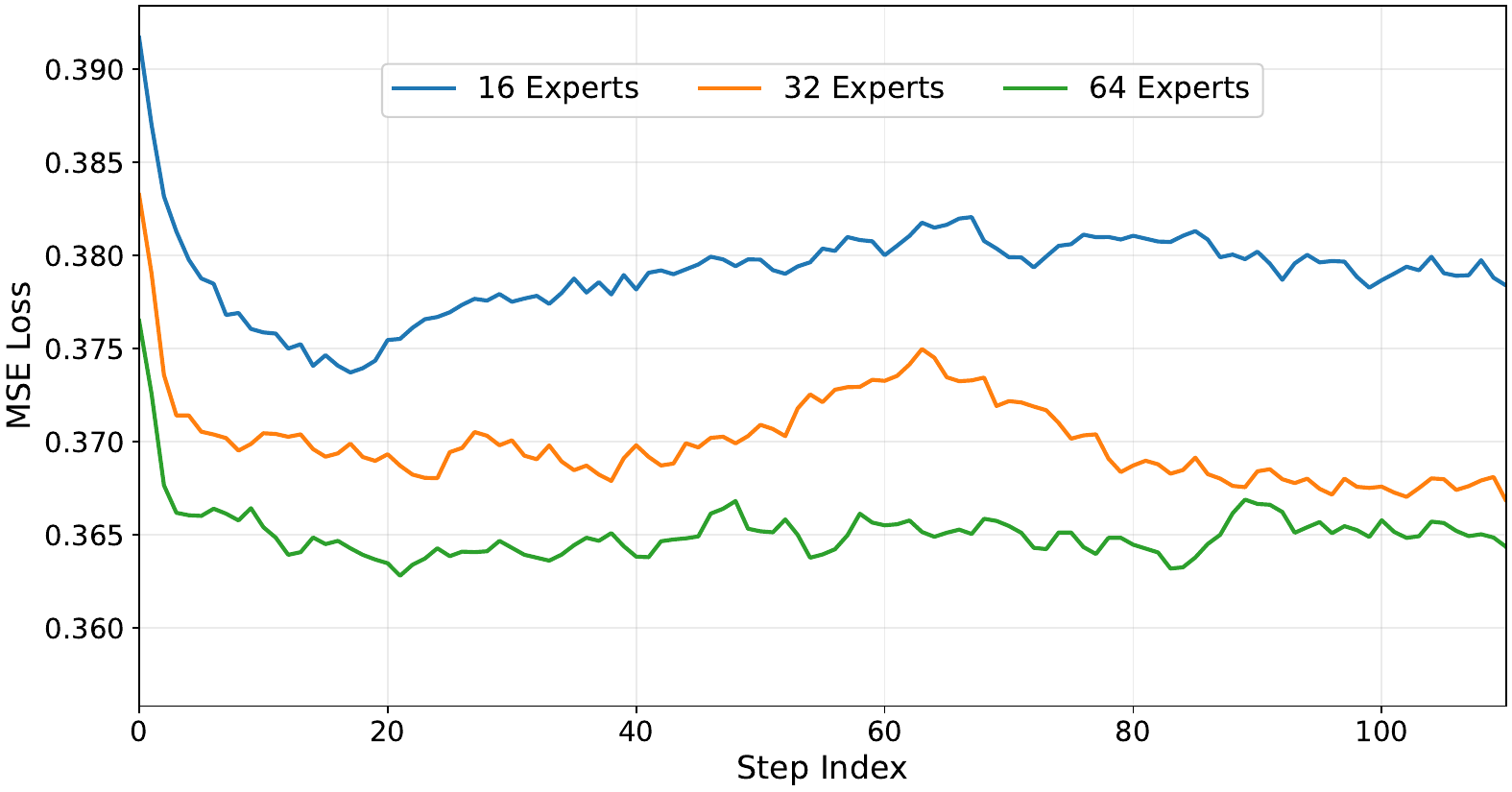} 
  \vspace{-5pt}    
\caption{Training curves of expert-frozen tuning with only a few training steps, where MoE layers are configured with different numbers of experts (i.e., 16, 32 and 64). }
\label{fig:rt_loss}
  \vspace{-12pt}   
\end{figure}

\paragraph{\textbf{Warmup via Expert-Frozen Tuning}}

MoE Adaptation begins with Expert-Frozen Tuning, which serves as a cold start phase to train the model to effectively leverage the partitioned experts. This strategy mitigates catastrophic forgetting and encourages the model to learn effective expert selection while preserving pretrained knowledge \cite{houlsby2019parameter, qiao2024learn, he2025routertuningsimpleeffectiveapproach}. Specifically, we examine different numbers of experts by comparing three configurations (16, 32, and 64), as shown in Figure~\ref{fig:rt_loss} depicting the training loss curves. With a few steps of expert-frozen tuning, we observe a substantial decline in loss values, indicating that the model effectively adapt to and exploit a subset of experts to recover performance. Meanwhile, finer-grained expert partitioning enables more flexible activation combinations, leading to substantially lower training loss.

\begin{figure*}[h]
  \centering
  \setlength{\tabcolsep}{2pt}
  \renewcommand{\arraystretch}{1.2} 
\resizebox{0.98\linewidth}{!}{
\begin{tabularx}{\textwidth}{
>{\raggedright\arraybackslash}m{0.2\textwidth} |
>{\centering\arraybackslash}m{0.15\textwidth} |
>{\centering\arraybackslash}m{0.15\textwidth} |
>{\centering\arraybackslash}m{0.15\textwidth} |
>{\centering\arraybackslash}m{0.15\textwidth} |
>{\centering\arraybackslash}m{0.15\textwidth}
} 
  
    \toprule
\textbf{Prompt} & 
\textbf{Baseline} & 
\textbf{Zeroshot w/o SE} & 
\textbf{Zeroshot w/ SE} & 
\textbf{E.F. Tuning} & 
\textbf{MoE Adapt.} \\

\midrule
    \parbox[c]{0.2\textwidth}{\raggedright A famous flower that symbolizes wealth in China. } 
    &
\cellimg{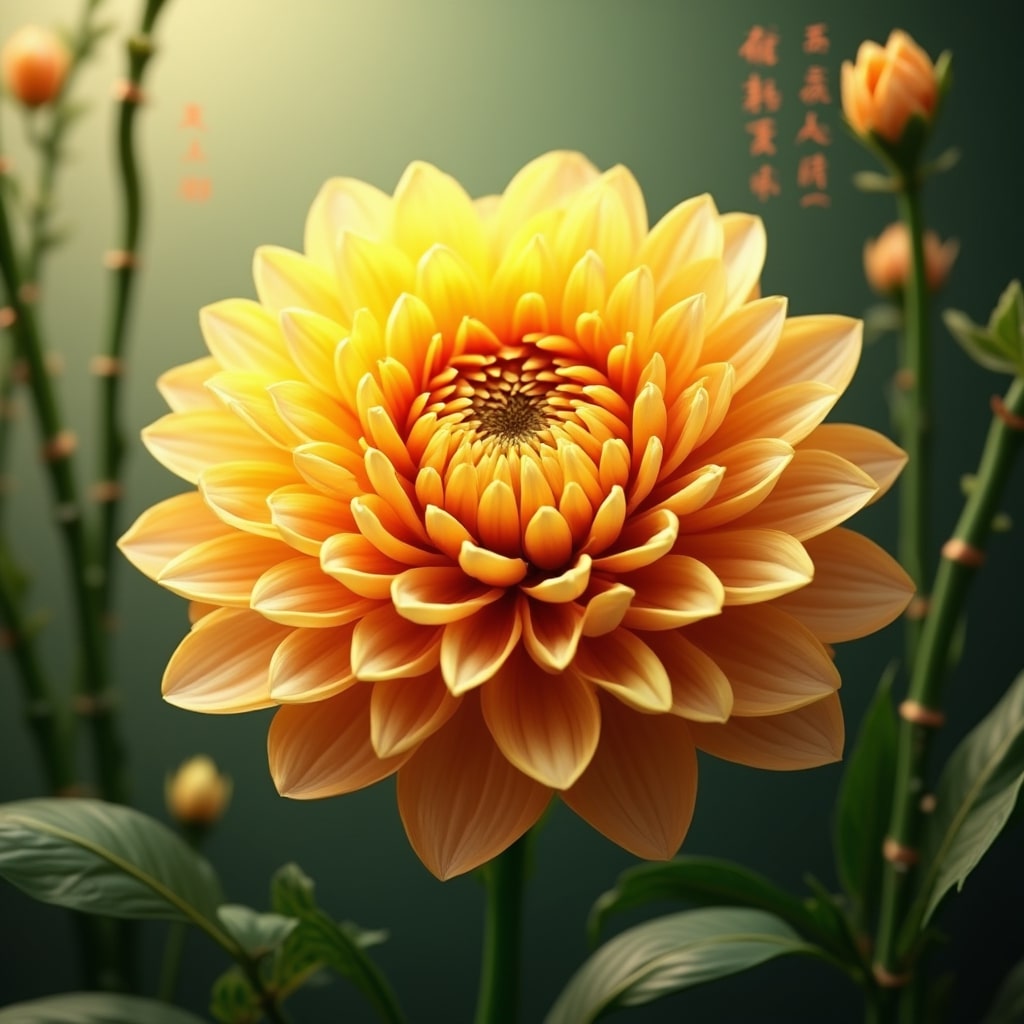} &
    \cellimg{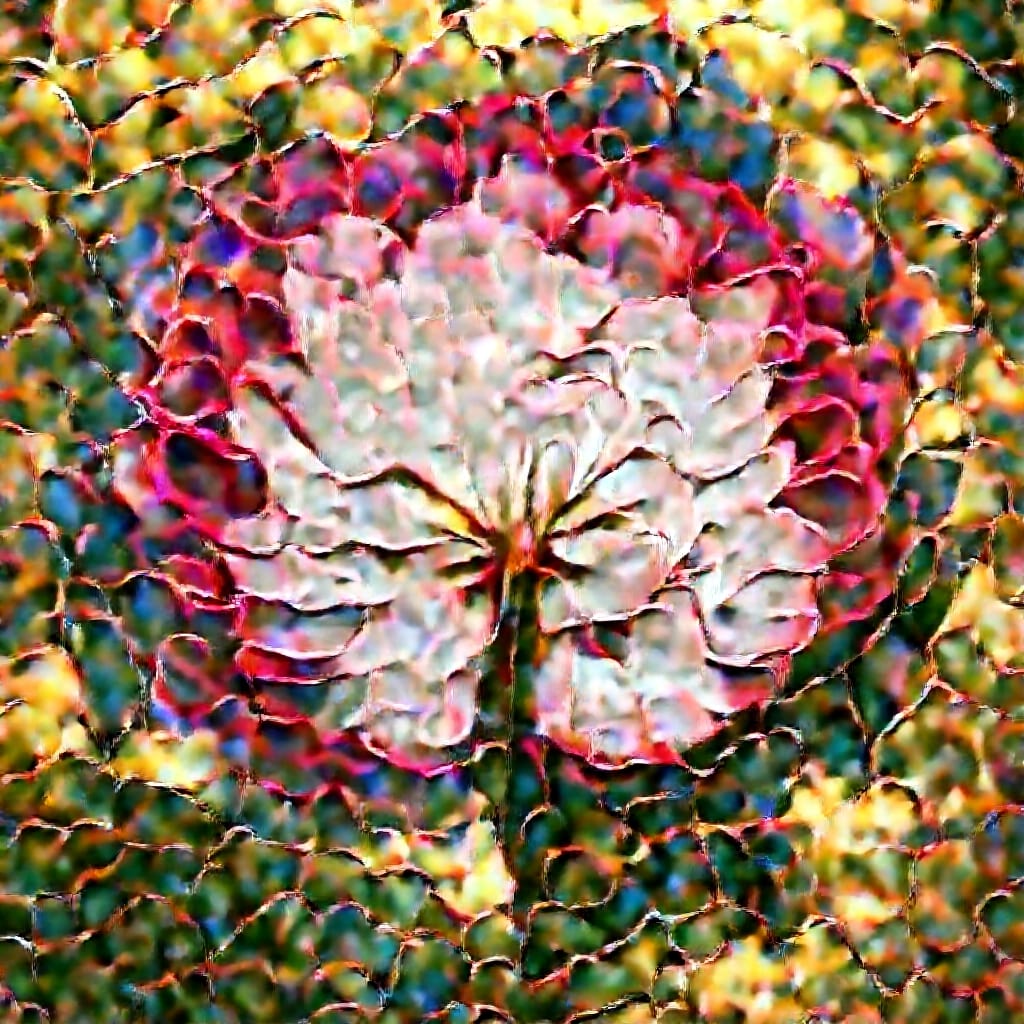} &
    \cellimg{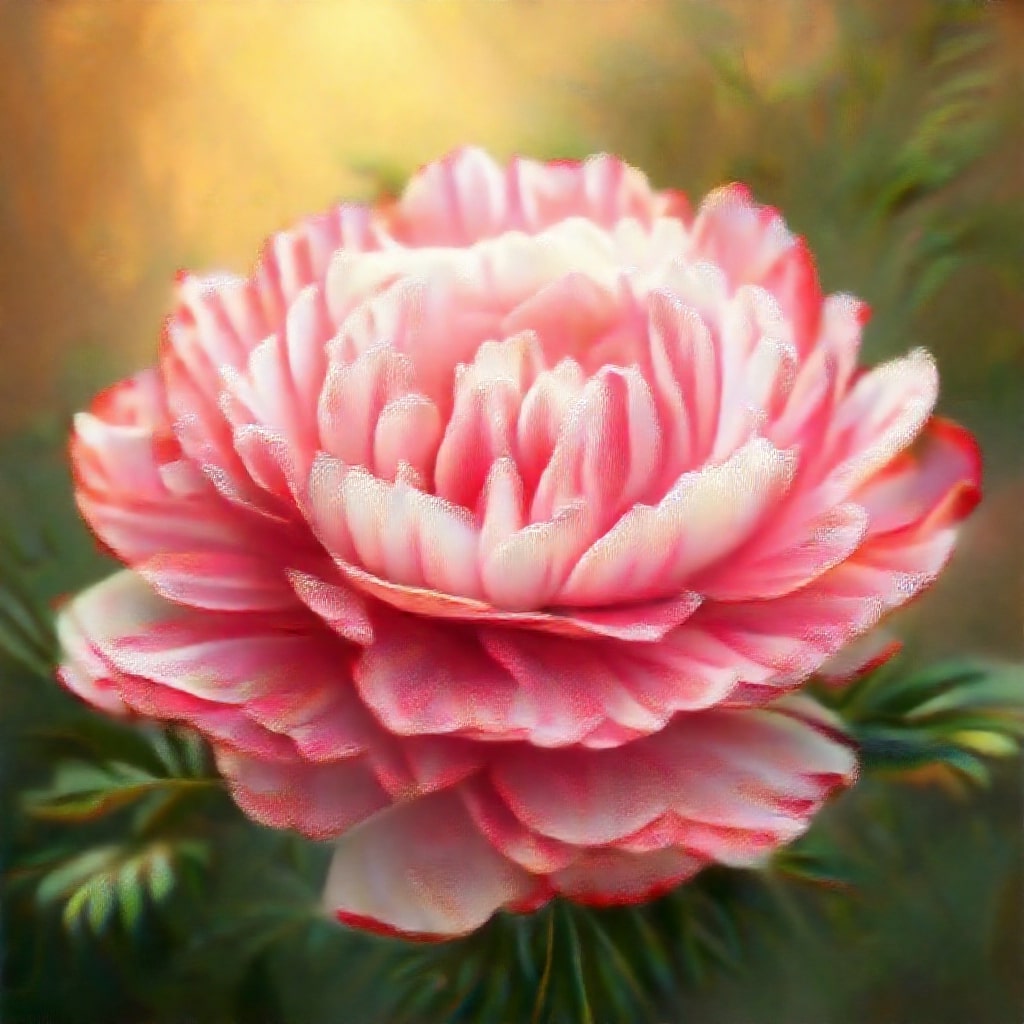} &
    \cellimg{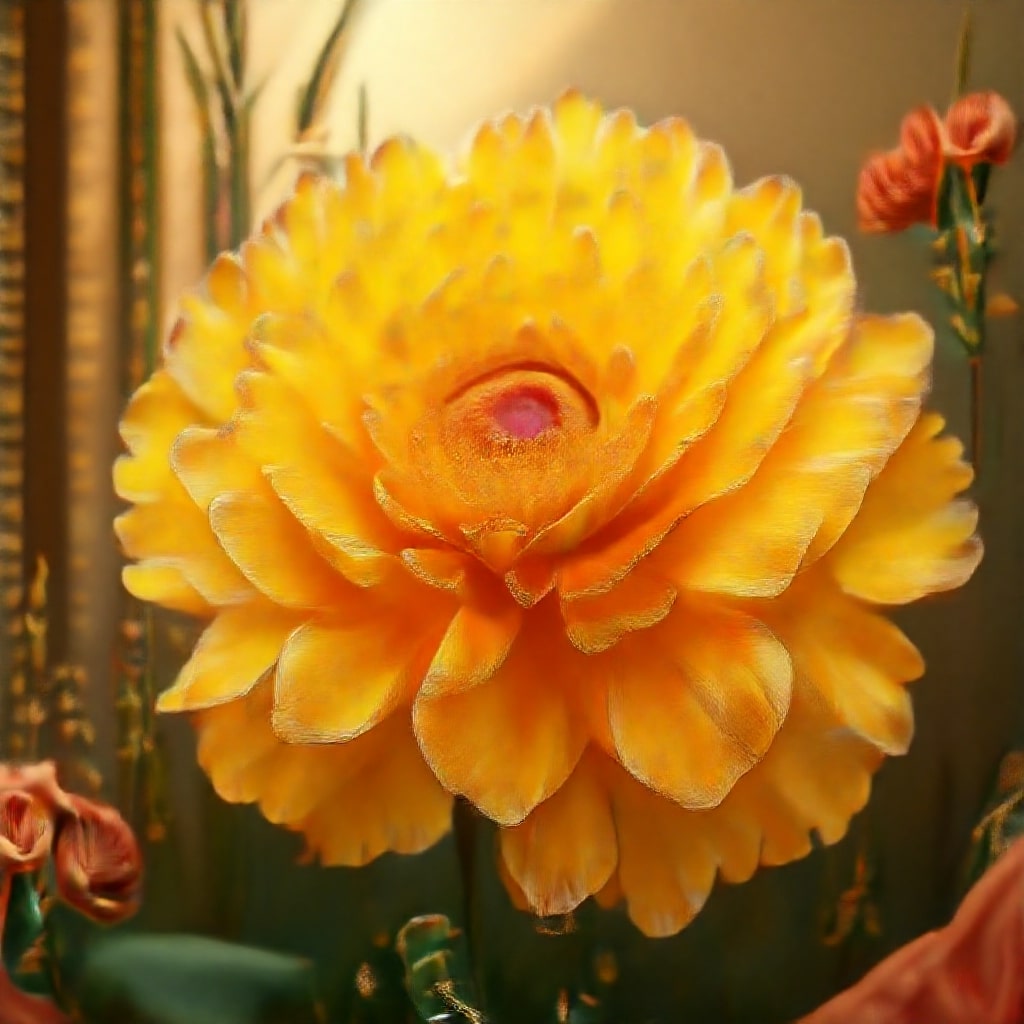} &
    \cellimg{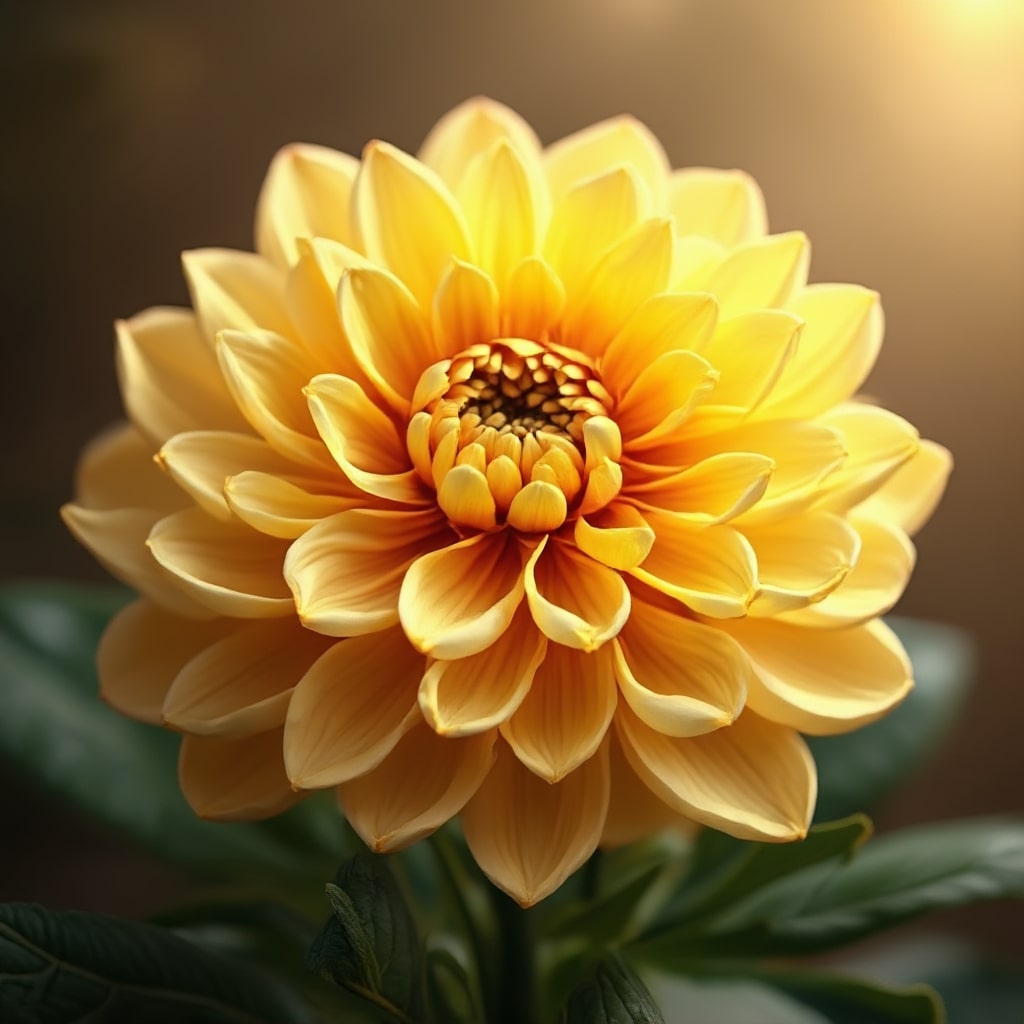} \\

\midrule
    \parbox[c]{0.2\textwidth}{\raggedright Traditional activity during Easter in Western countries. } &
\cellimg{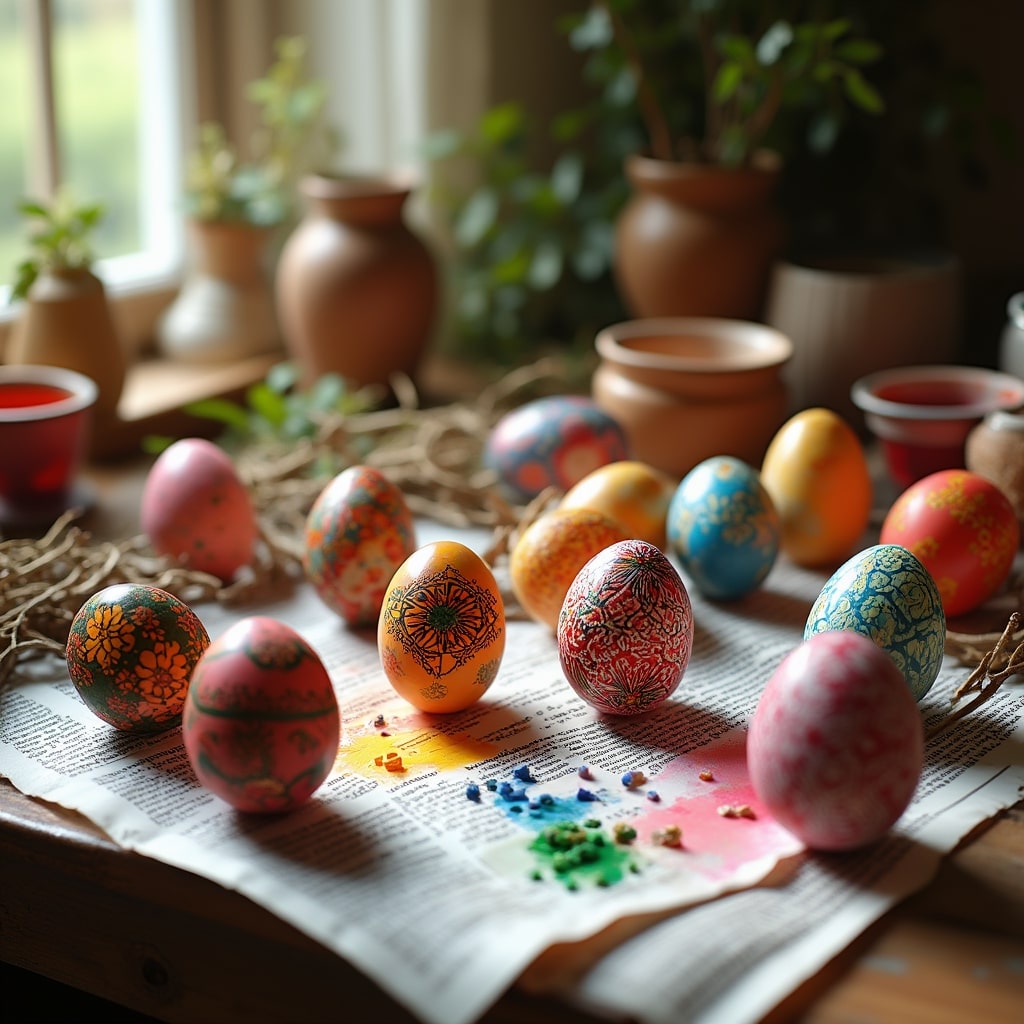} &
    \cellimg{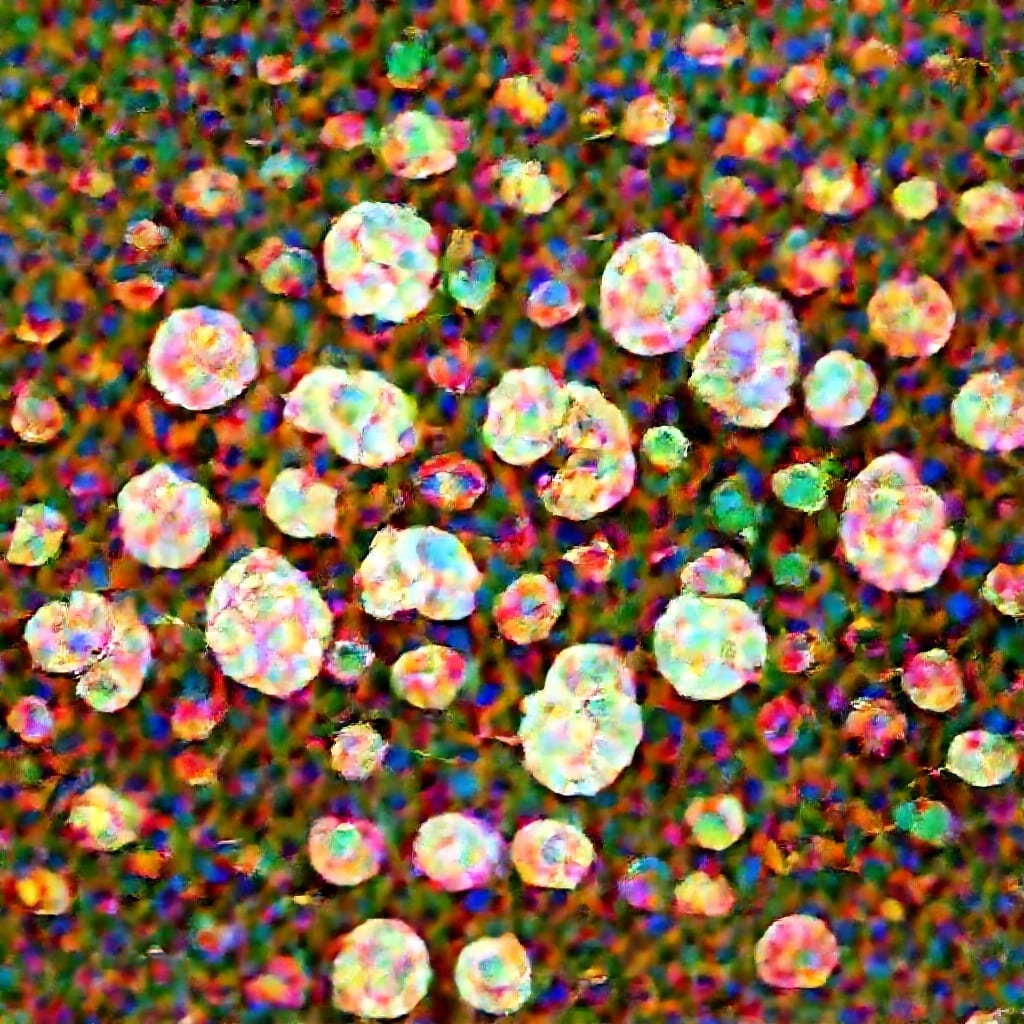} &
    \cellimg{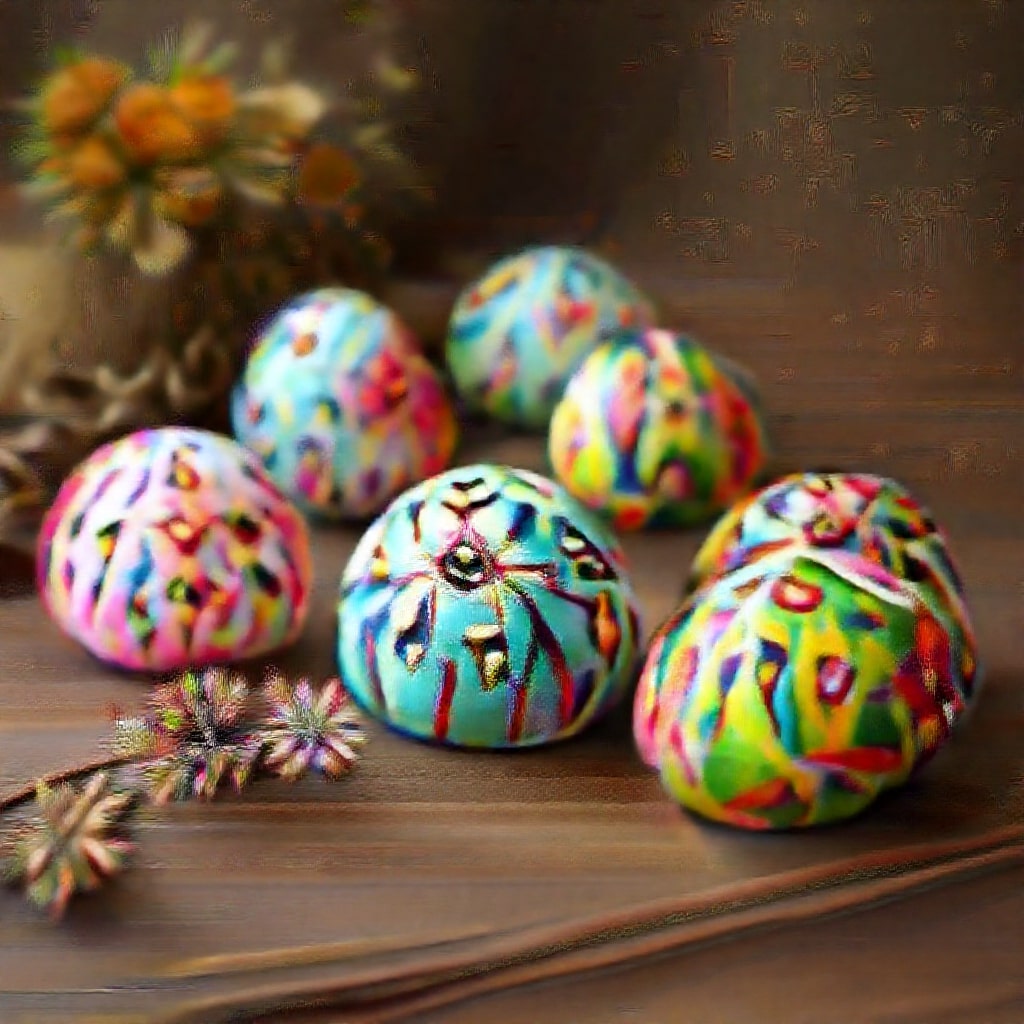} &
    \cellimg{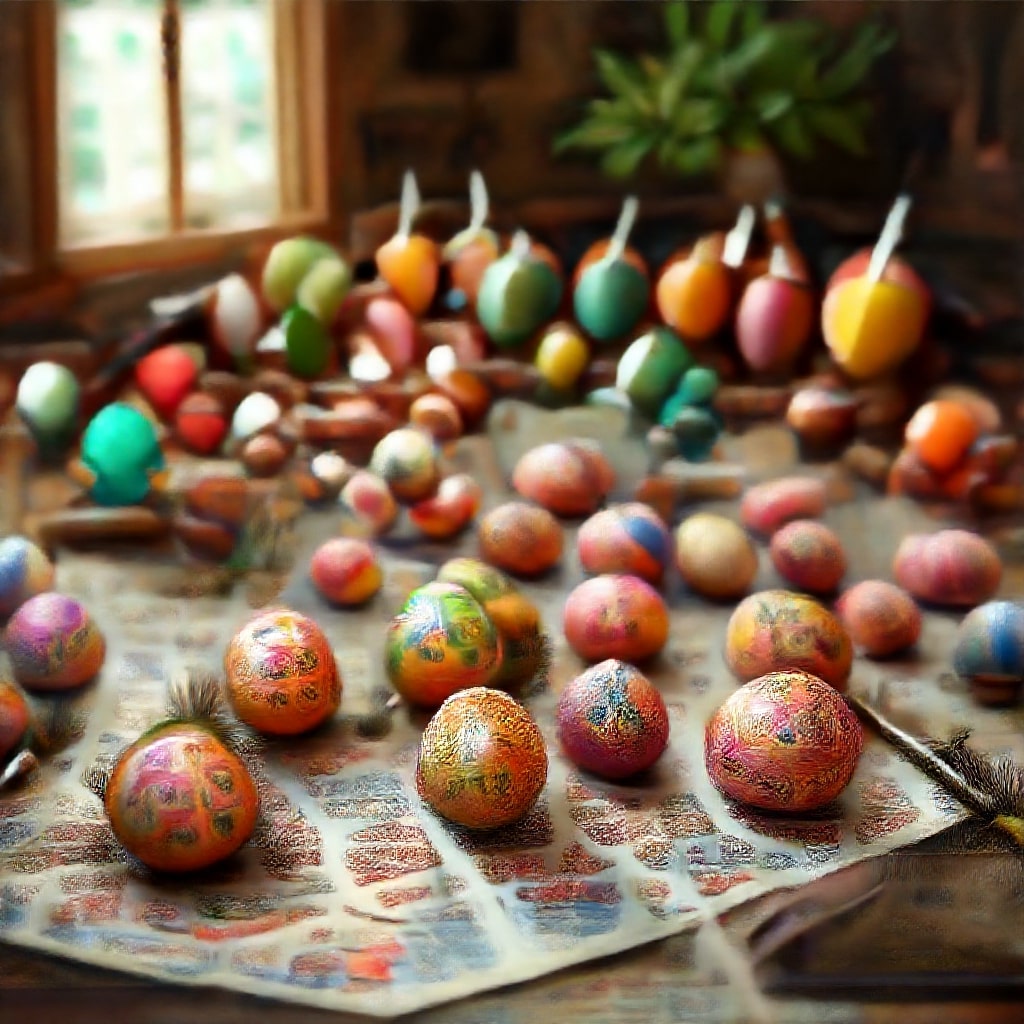} &
    \cellimg{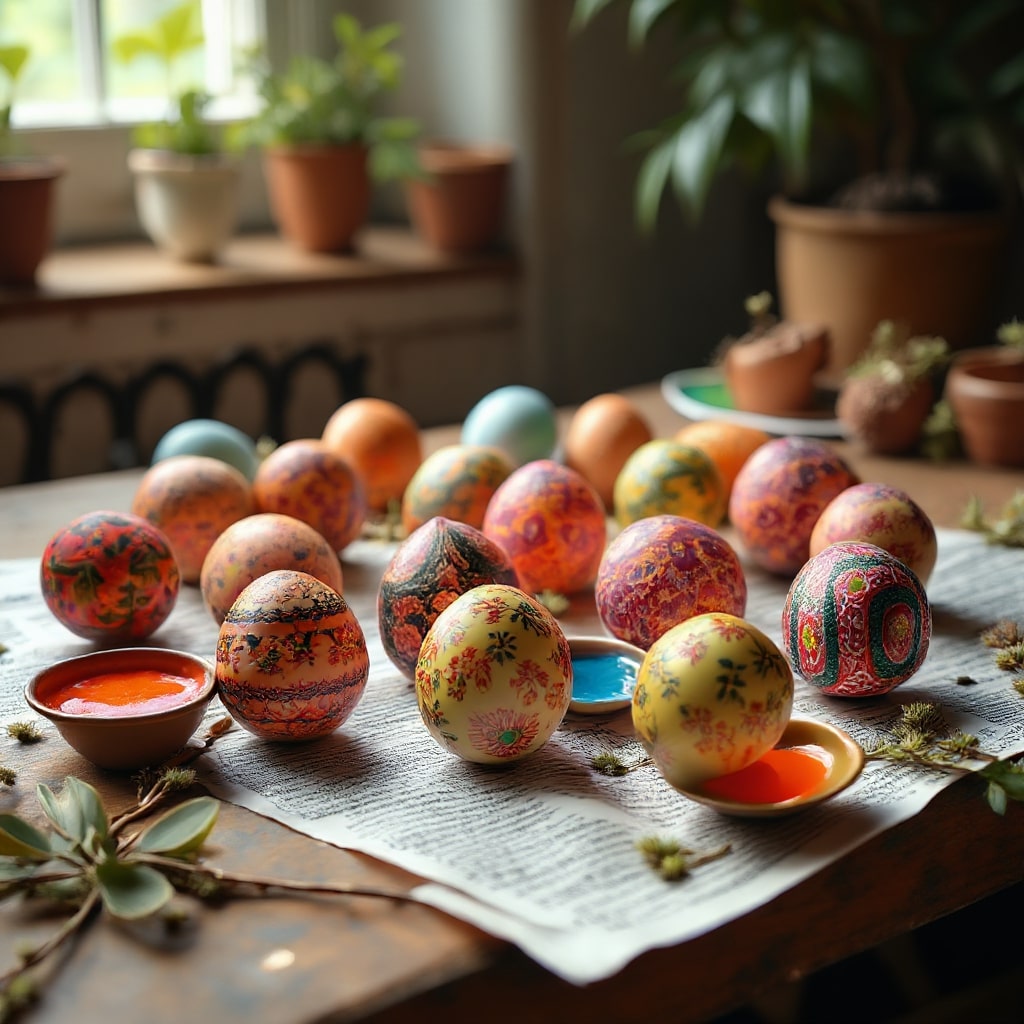} \\

\midrule
    \parbox[c]{0.2\textwidth}{\raggedright Old analog picture of parked car on side street, quiet night.  } &
\cellimg{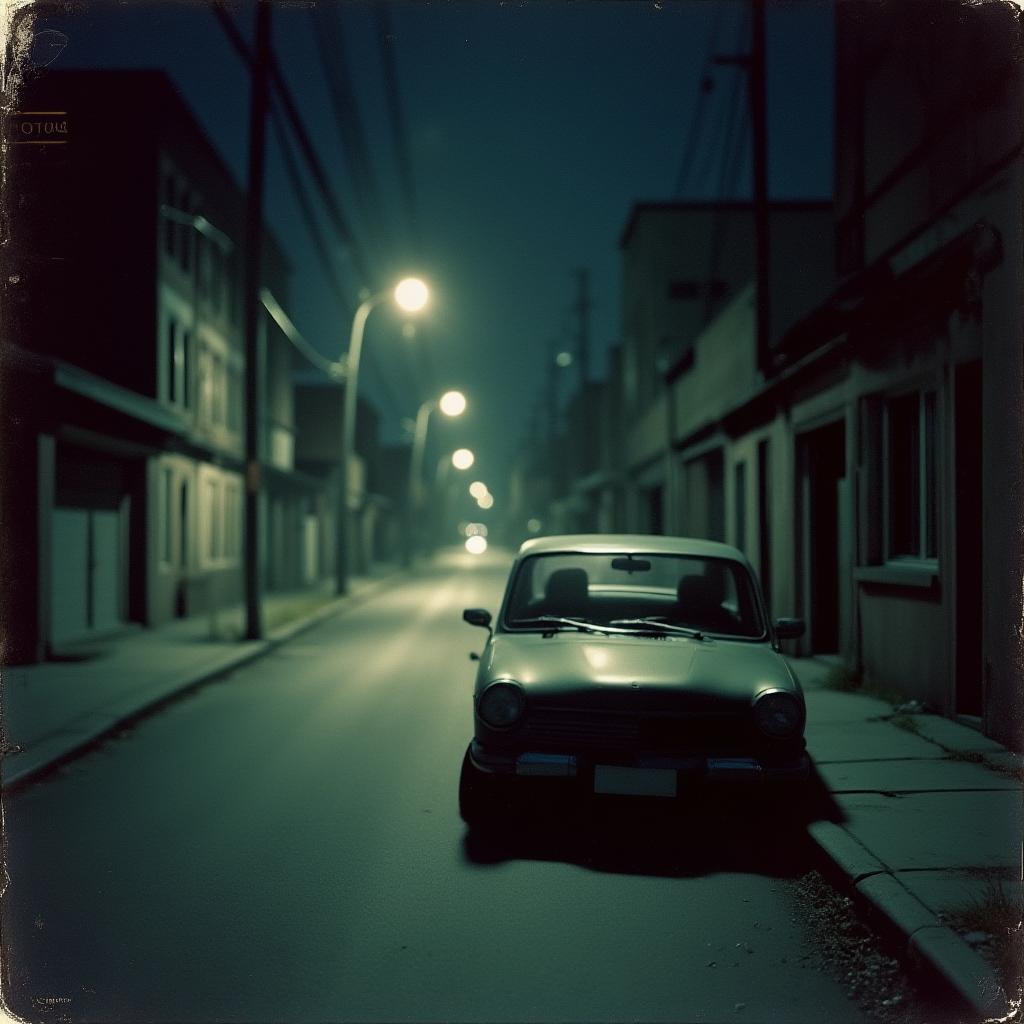} &
    \cellimg{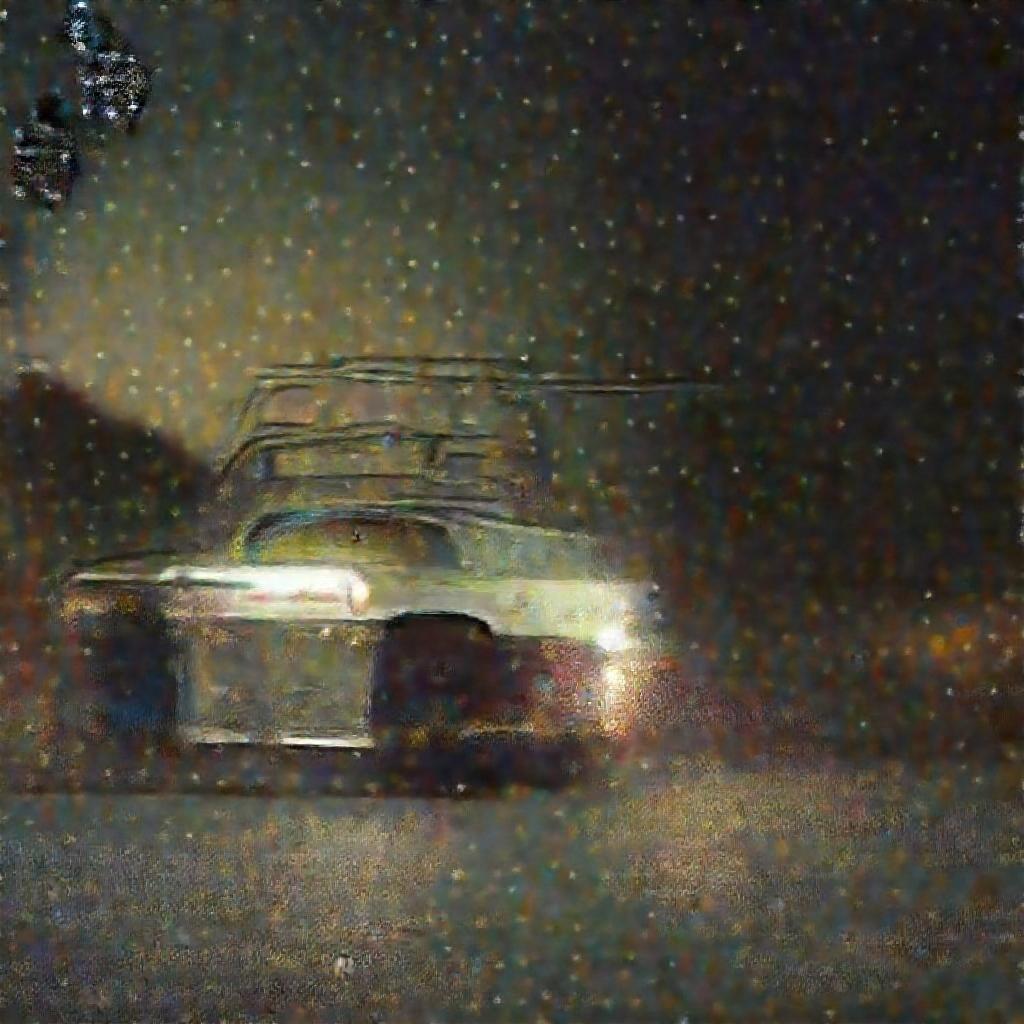} &
    \cellimg{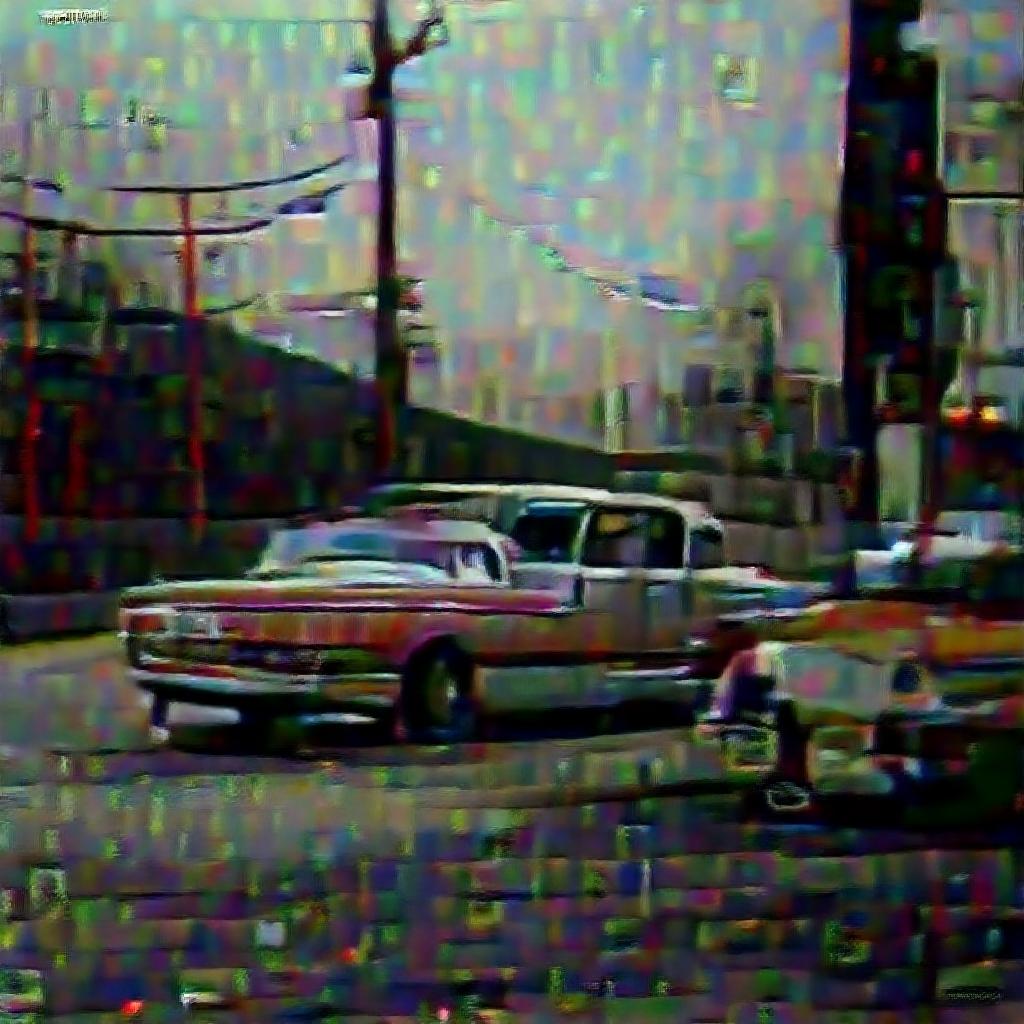} &
    \cellimg{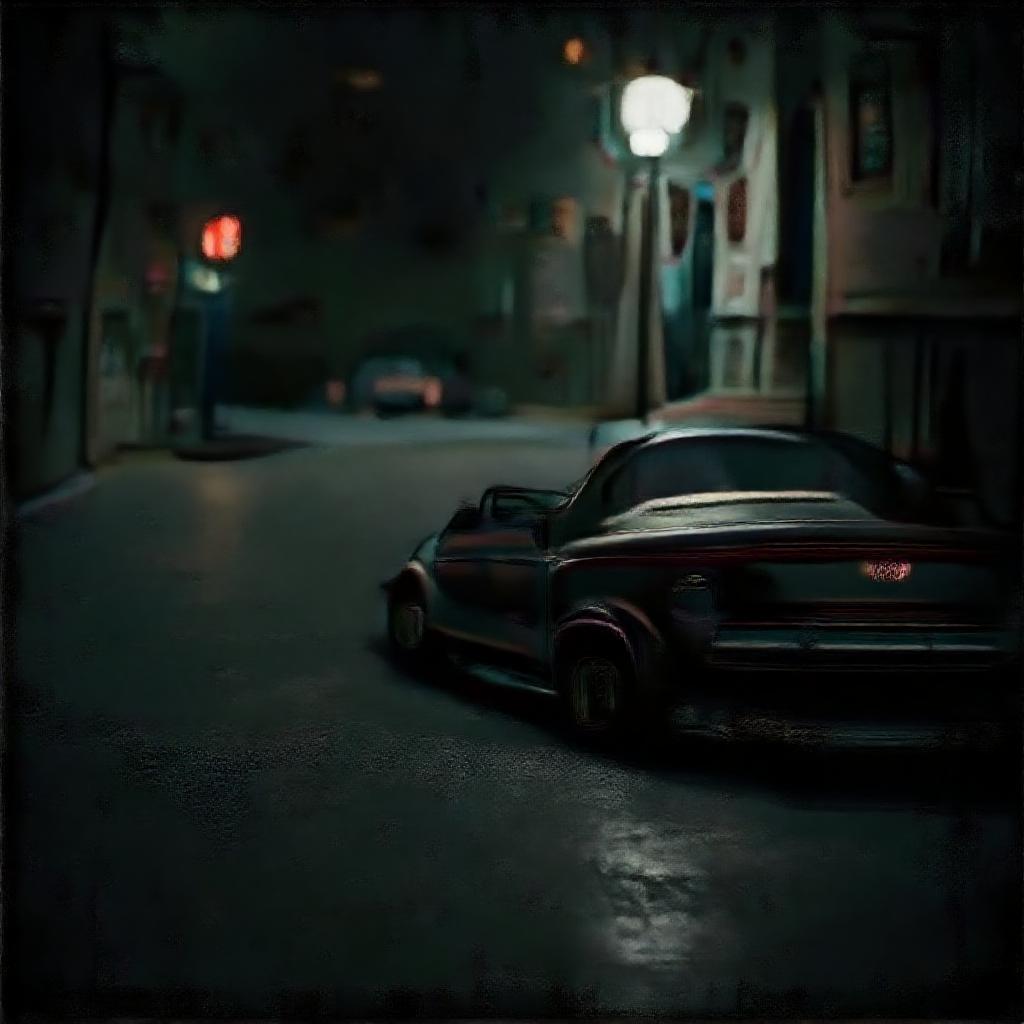} &
\cellimg{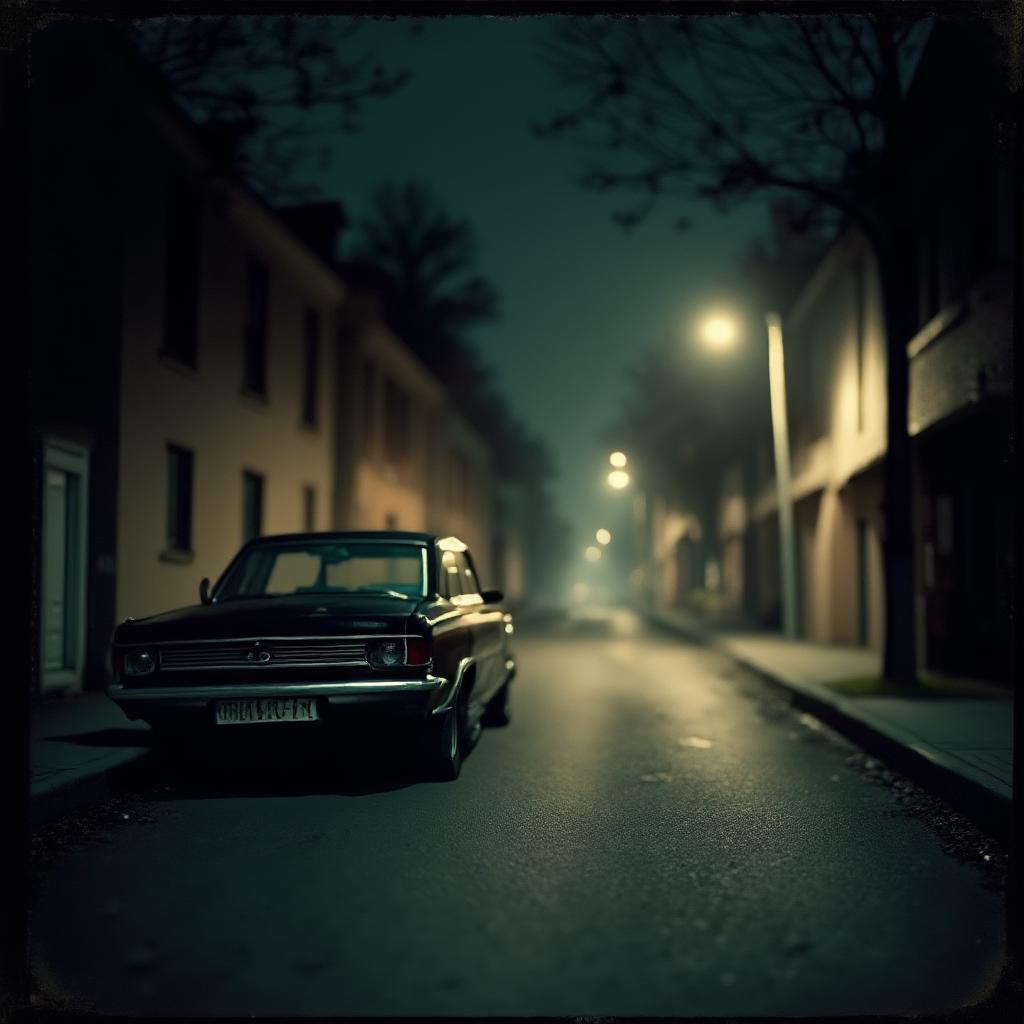} \\
    \midrule
\parbox[c]{0.2\textwidth}{\vspace{-10pt}\raggedright A lone astronaut paints swirling galaxies onto a massive canvas in a vast space station. } &
\cellimg{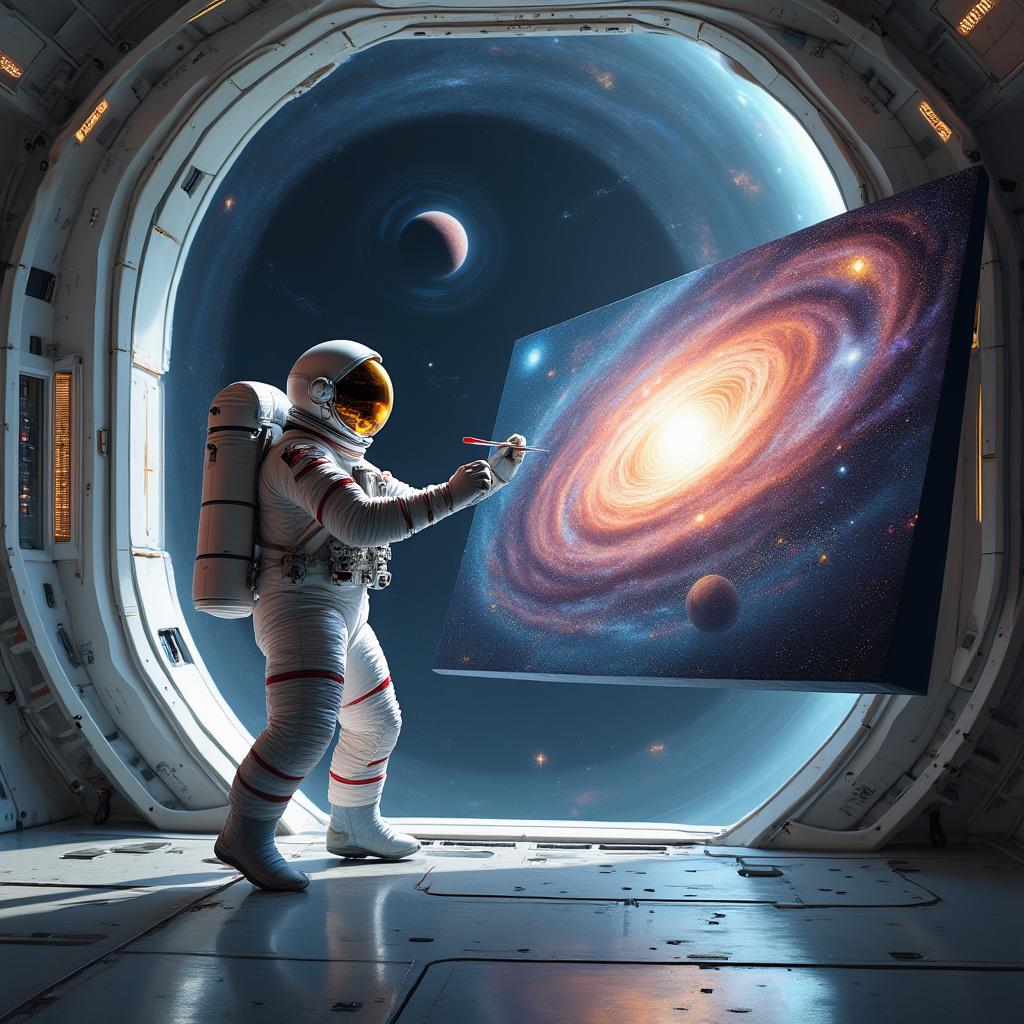} &
    \cellimg{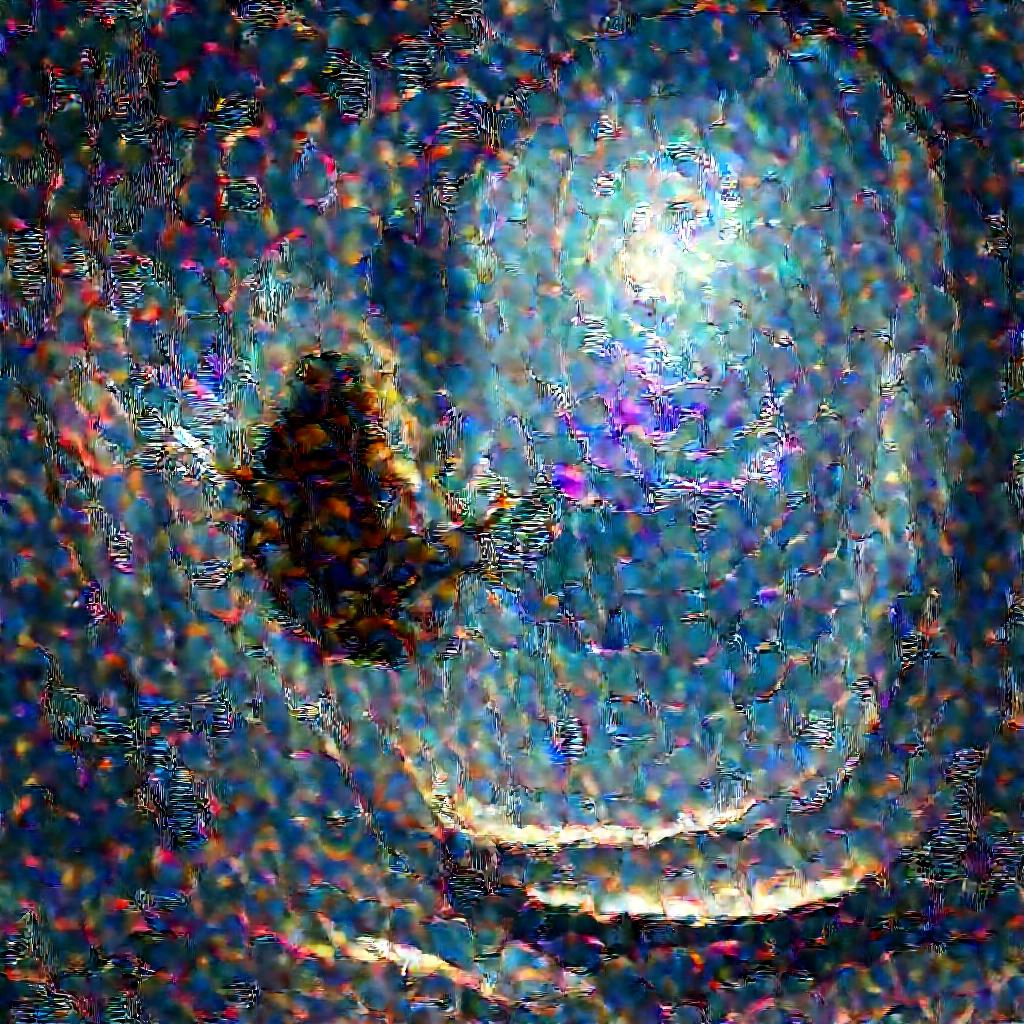} &
    \cellimg{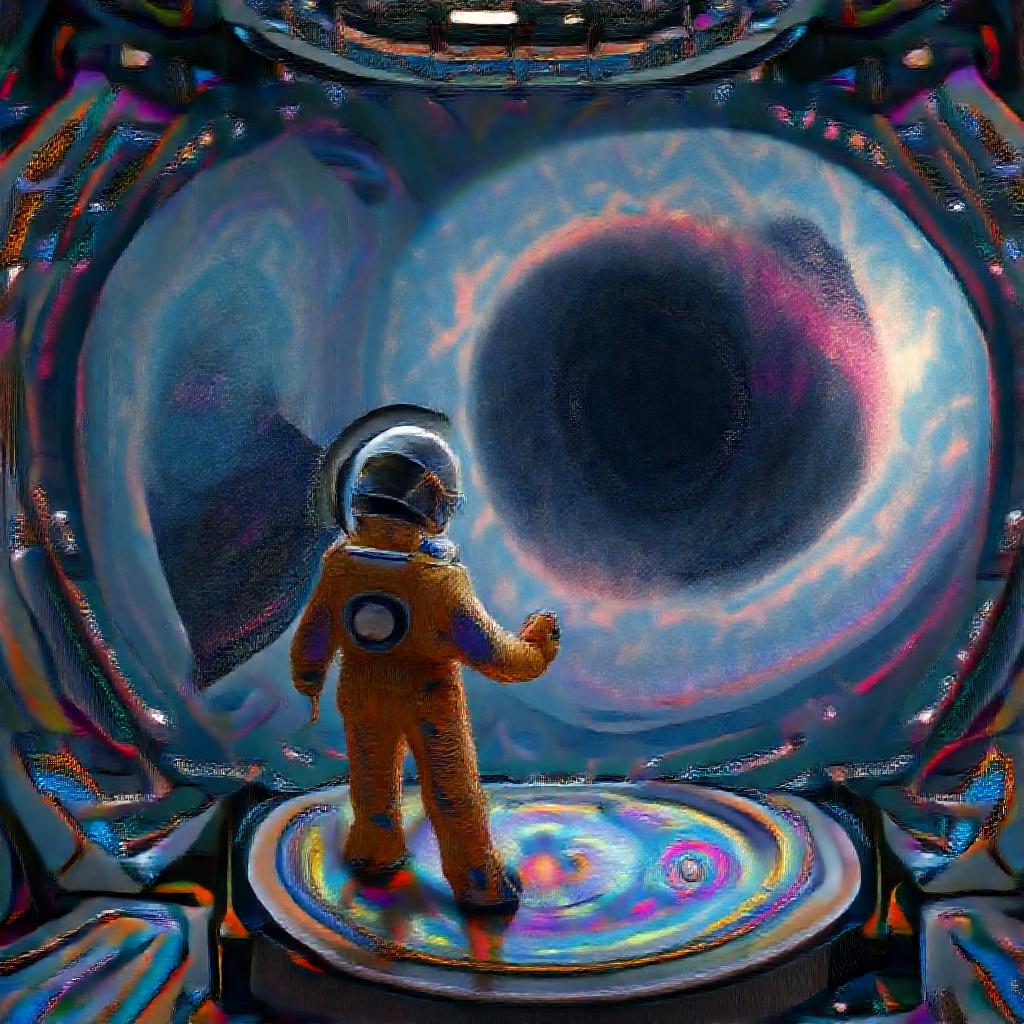} &
    \cellimg{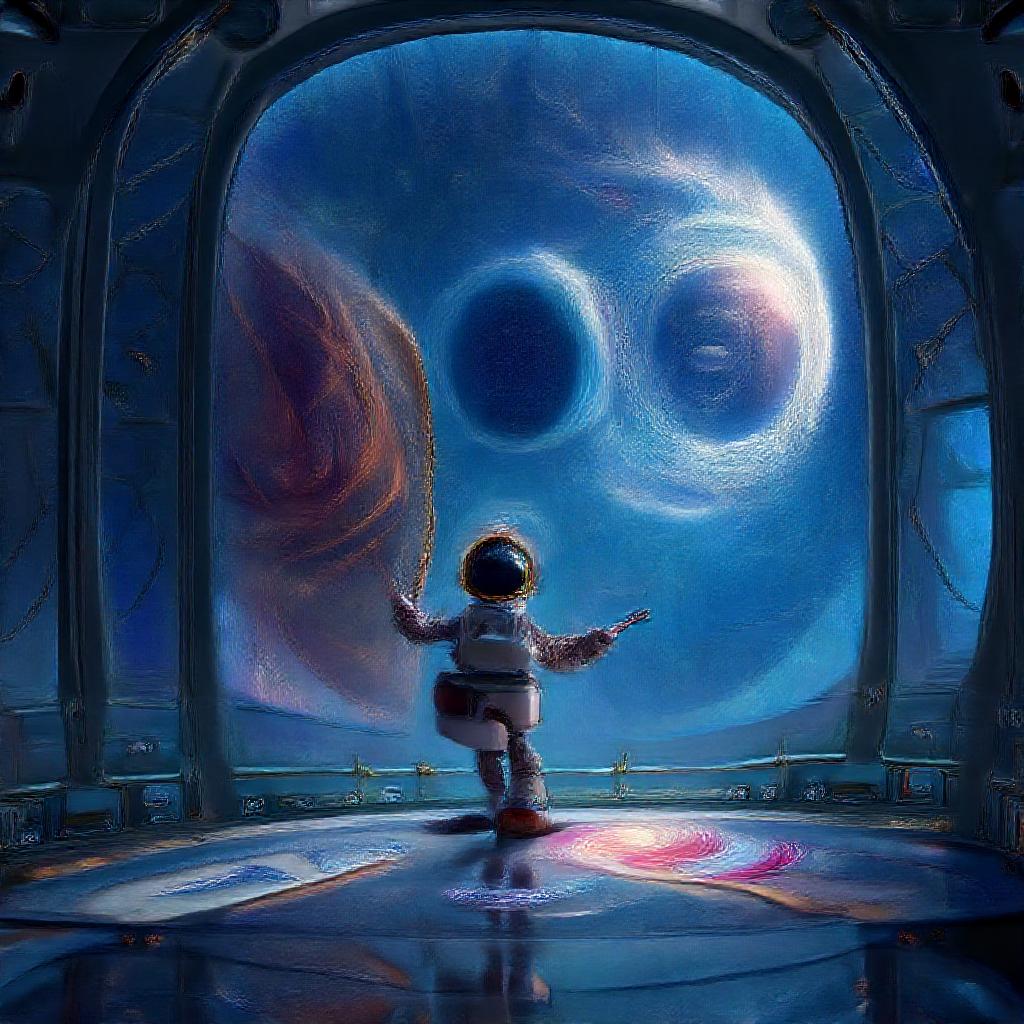} &
    \cellimg{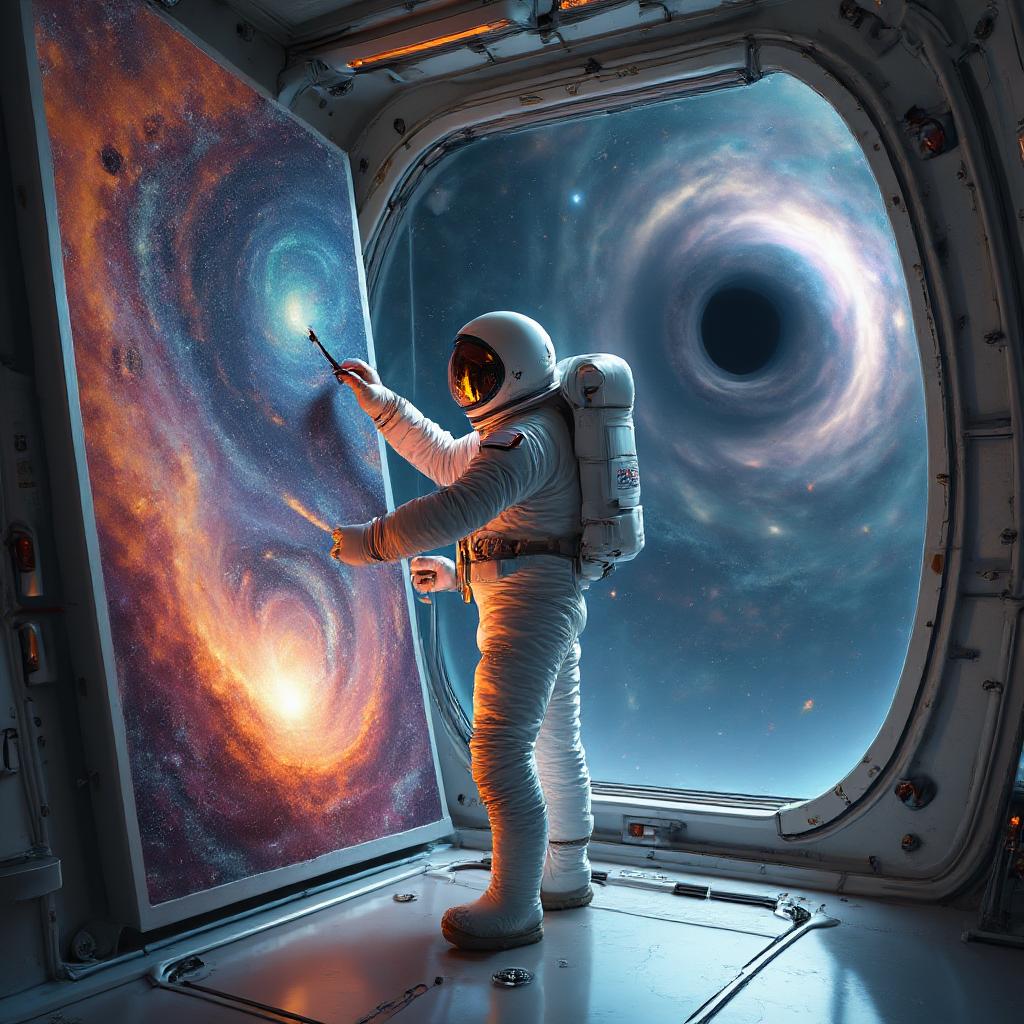} 
    \\ 

    \bottomrule
  \end{tabularx}}
  \vspace{5pt}
\caption{
\textbf{Visual comparison across different stages of MoE Adaptation.}
Shown are outputs from the \textbf{Baseline} model (without modification), \textbf{Zeroshot} settings after expert partition with or without shared experts (\textbf{Zeroshot w/o S} and \textbf{Zeroshot w/ S}), the model after \textbf{Expert-Frozen Tuning (E.F. Tuning)}, and the model further optimized via \textbf{MoE Adaptation (MoE Adapt.)}.
Test prompts are sampled from WISE~\citep{niu2025wise} and the 4o-Image Generator prompt set. 
}
  \label{fig:training-comparison}
\end{figure*}

For instance, the overall GenEval score improves from 0.58 to 0.78, reflecting more coherent and visually faithful outputs. As illustrated in Figure~\ref{fig:training-comparison}, before tuning, the model generates noisy, low-detail images that fail to capture fine-grained semantics. Expert-Frozen Tuning not only enhances image fidelity but also strengthens the alignment between the generated content and the given instructions. This demonstrates that certain subsets of parameters within the generation components, though difficult to compress, still retain the potential to produce high-quality images.

\paragraph{\textbf{Effectiveness of MoE Adaptation}}

After a few steps of Expert-Frozen Tuning as a cold start, we lift the restriction on frozen expert parameters to further improve performance.
Beyond applying MoE adaptation exclusively to the generation component, we also explore extending it to the understanding component to reduce the number of activated parameters while maintaining task effectiveness.
To preserve fidelity in understanding tasks, the experts in the understanding component are kept frozen, remaining fully activated for understanding and only sparsely activated for generation, since generation tasks are more tolerant to sparsity in this component. In summary, we consider two configurations of MoE adaptation:
(1) \textit{Gen.}: Expert partitioning and adaptation are applied only to generation experts; and
(2) \textit{Und. \& Gen.}: Both understanding and generation experts are partitioned, but only the generation experts are adapted while the understanding experts remain frozen. 

These additional training stages enable the experts to refine their internal representations and develop stronger specialization, improving both structural coherence and semantic consistency in generated outputs. As shown in Table~\ref{tab:router-tuning} and Figure~\ref{fig:training-comparison}, the model achieves consistently higher generation quality after Expert-Frozen Tuning and further gains after full MoE Adaptation. Although generation components are more sensitive to training-free compression, sparsely activating fewer parameters still has the potential to preserve the original performance. 

\section{Conclusion}
This work investigates the slimness and sparsity of unified multimodal models, proposing training-free pruning and dynamic activation methods to enhance efficiency.
Our investigation reveals that the understanding component exhibits strong compressibility across both understanding and generation tasks.
In contrast, the generation component is far more sensitive to compression, where even moderate pruning can lead to catastrophic degradation in generation quality.
Inspired by the dynamic activation patterns observed across different samples, we introduce a Mixture-of-Experts (MoE) Adaptation that dynamically activates generation parameters, effectively restoring generation quality under sparse activation.
By combining training-free analysis with training-aware adaptation, our study uncovers substantial optimization space for improving parameter efficiency in unified multimodal modeling. 

\section{Acknowledgment}
We sincerely thank Shuangye Li, Yuhong Yang, and Yi Lin for their technical discussion that contributed to the development of this work.

{
    \small
    \bibliographystyle{ieeenat_fullname}
    \bibliography{main}
}

\clearpage 

\appendix

\section{Generation Component Compression}

\label{app:gen_comp}

\begin{figure}[h]
  \centering
  \includegraphics[width=\linewidth]{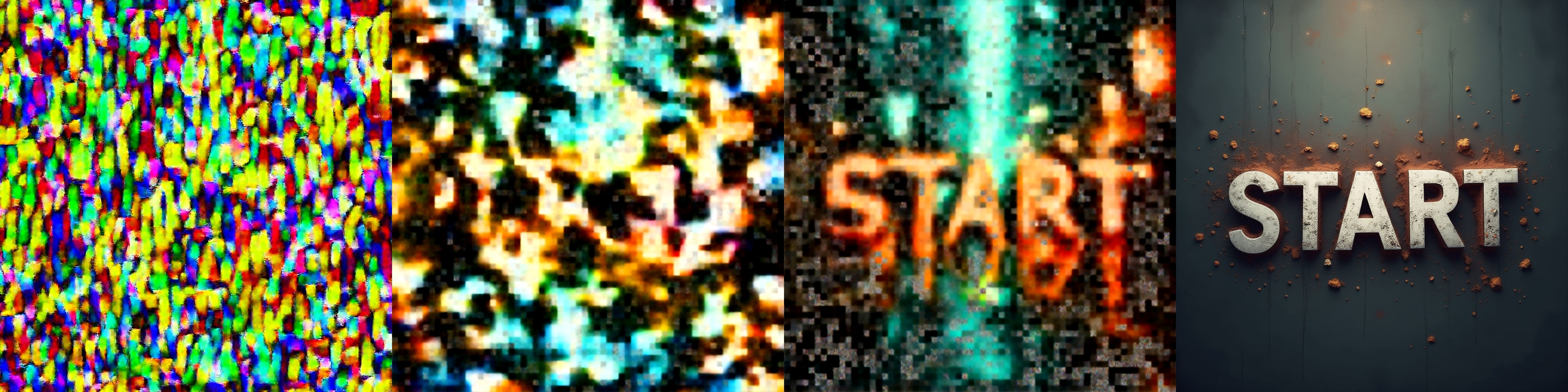} 
\caption{\textbf{Depth reduction applied to MLP layers in the generation component. }
Figures are shown with decreasing numbers of removed layers: 14 (50\%), 7 (25\%), 4 (14\%), and 0.}   \label{fig:depth_reduction_gen}
\end{figure}

Generation components exhibit substantially higher sensitivity to compression than understanding components. Beyond the neuron partition results, we further investigate depth reduction (Figure~\ref{fig:depth_reduction_gen}) and find that removing entire layers also causes catastrophic degradation in generated outputs. These results indicate that, for generation components, static compression along either width or depth alone struggles to preserve the performance of the original model. 

\begin{figure}[ht]
\centering

\begin{minipage}[t]{0.48\textwidth}
    \centering
    \includegraphics[width=\textwidth]{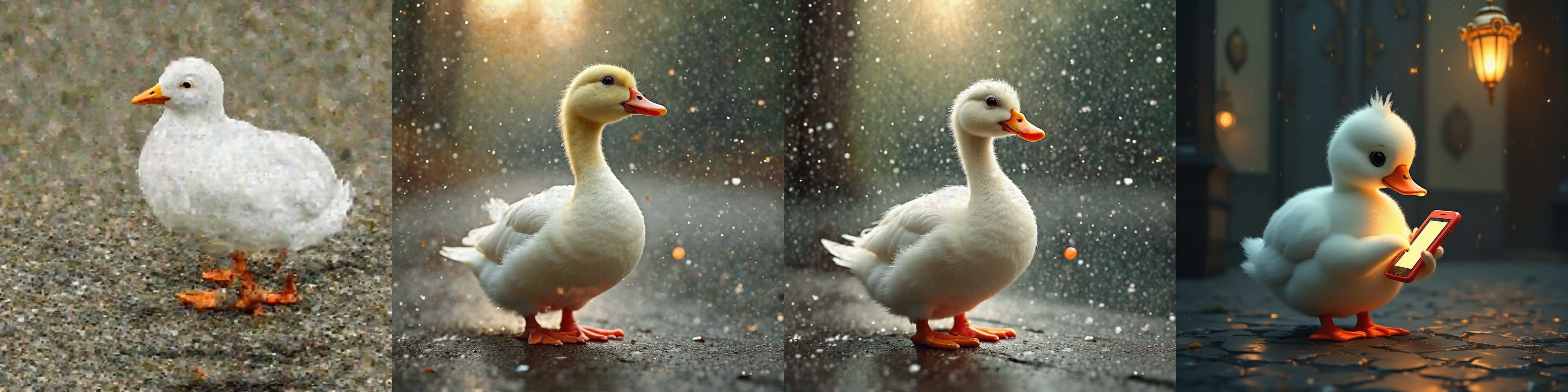}
    \parbox[t]{\textwidth}{\small 
    (a) 
    Depth reduction achieved by removing 7, 4, 2, or 0 layers.}
\end{minipage}
\hfill
\begin{minipage}[t]{0.48\textwidth}
    \centering
\includegraphics[width=\textwidth]{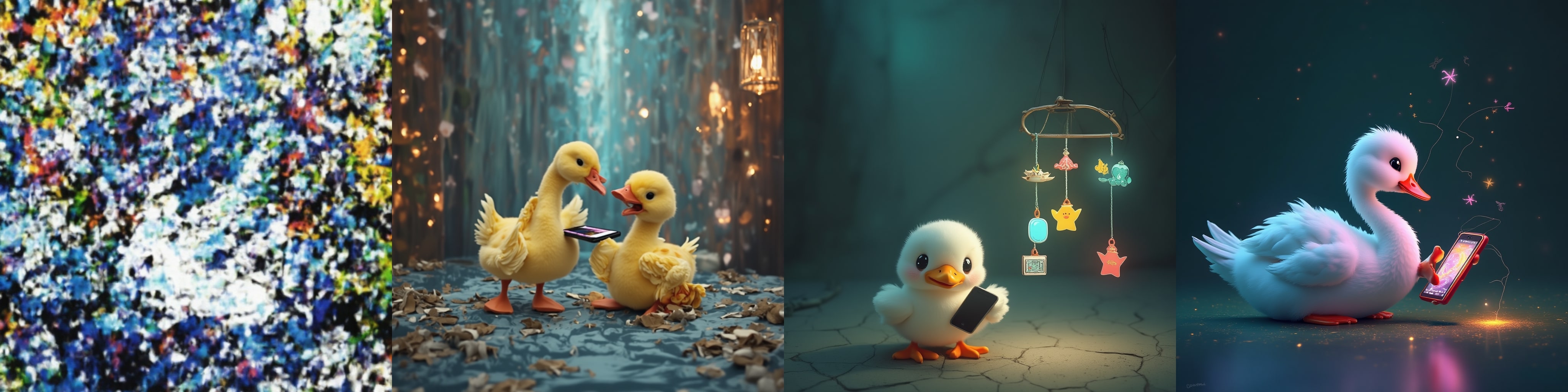}\\
    \parbox[t]{\textwidth}{\small (b) Width reduction evaluated at ratios of 50\%, 25\%, 10\%, and 0.}
\end{minipage}
\caption{Compression of generation components through pruning of attention layers and heads.}
\label{fig:attn_gen_comp}
\end{figure}

On the other hand, compressing the attention layers leads to substantial degradation in both depth and width settings. As shown in Figure~\ref{fig:attn_gen_comp}, applying more than a 10\% reduction results in noticeable performance drops.

\section{Attention Head Partition}
\label{app:attn_heads}

While our main focus is width reduction in MLP layers, since they account for most of the parameters, our compression metric can be seamlessly extended to attention by computing the importance scores of attention heads. Attention Head Parttion still works for understanding components as shown in Table \ref{tab:width_reduction_attn}.

\begin{figure*}[!t]
\centering
\begin{minipage}[c]{0.45\textwidth}
    \centering
\includegraphics[width=\linewidth]{figs/meme.jpg}
    \vspace{-25pt}
    \caption*{\small Meme: ``My Handwriting In Exams''}
\end{minipage}%
\hfill
\begin{minipage}[c]{0.50\textwidth}
    \raggedright
    \textbf{Question:} Can someone explain what’s funny about this meme? \\[0.8em]

    \textbf{Neuron Partition: } \\
    The humor in this meme comes from the exaggerated change in handwriting style as the exam progresses. At the beginning, the handwriting is clear and legible, indicating a confident and well-prepared student. However, as the exam continues, the handwriting becomes increasingly difficult to read, suggesting that the student is becoming more anxious and less focused. \\[0.8em]

    \textbf{Depth Reduction: } \\
    The comedic resurgence biting meme portrays a comedic portrayal portraying a trope portraying idioms portraying idioms portraying idioms portraying idioms portraying idioms portraying idioms portraying exposes exposing exposing injuring injuring injuring injuring injuring injuring injuring injuring injuring injuring......
\end{minipage}
\caption{\textbf{Comparison of neuron partition and depth reduction on a toy example under a 50\% compression ratio. }
The model with fewer neurons per layer still answers the question reasonably well, whereas the model with reduced depth collapses and degenerates into repeating a single word. } 
\label{fig:handwriting-meme}
\end{figure*}

\begin{table}[!t]
\centering
\caption{Performance of neuron partition applied on attention heads at a sparsity ratio of 50\% per layer.}
\vspace{-6pt}
\resizebox{\linewidth}{!}{
\begin{tabular}{l|c|cccccc|c}
\toprule
\textbf{Model} 
& \textbf{Compressed Layers} & \textbf{Single Obj.} & \textbf{Two Obj.} & \textbf{Counting} & \textbf{Colors} & \textbf{Position} & \textbf{Color Attri.} & \textbf{Overall}$\uparrow$ \\
\toprule
\multirow{3}{*}{{BAGEL}}  
& N/A & 0.99 & 0.94 & 0.81 & 0.95 & 0.72 & 0.77 & \underline{0.86} \\
& 3-27 & 0.97 & 0.87 & 0.66 & 0.88 & 0.33 & 0.31 &  \underline{0.67}
 \\
& 4-27 & 0.98 & 0.91 & 0.72 & 0.89 & 0.41 & 0.40 & \underline{0.72}
 \\
\bottomrule
\end{tabular}}
\label{tab:width_reduction_attn}
\end{table}

\begin{table}[h]
\centering
\caption{Performance of depth reduction on understanding tasks. }
\vspace{-6pt}
\resizebox{\linewidth}{!}{
\begin{tabular}{l|c|ccccc}
\toprule
\textbf{Model} & \textbf{Sparsity} & \textbf{MME-P} & \textbf{MME-C} & \textbf{MMMU} & \textbf{MMBench} & \textbf{MMVP} \\
\midrule
\multirow{2}{*}{{Ming-Omni}}~~ & --   & 1584.3 & 670.4 & 66.7 & 86.7 & 54.6 \\
          & 50\% & 1197.2 & 308.2 & 51.7 & 81.2 & 46.0 \\
\midrule
\multirow{2}{*}{{BAGEL}}  
& --   & 1684.8 & 696.7 & 65.0 & 88.1 & 69.6 \\
          & 50\% &  304.5 & 127.1 & 16.7 & 18.6 & 23.1 \\
\bottomrule
\end{tabular}}
\label{tab:depth_reduction_und}
\end{table}

\section{Depth Reduction
 on Understanding Tasks}

While reducing depth in the understanding component has only a limited impact on generation tasks, it severely harms multimodal understanding performance, as shown in Table~\ref{tab:depth_reduction_und}. Moreover, Figure~\ref{fig:handwriting-meme} illustrates that the reduced-depth model fails to produce coherent answers, collapsing after only a few generated tokens. We attribute this to error accumulation in autoregressive decoding: small deviations in early generated tokens compound over time, ultimately causing the entire response to collapse.

\section{Comparison with Compression Methods}

While neuron partition leverages reconstruction errors to estimate the importance of each neuron, another mainstream approach relies on gradient information to approximate the impact of neuron removal. In Table \ref{tab:llmpruner}, we adopt gradient-based metrics, such as those used in LLM-Pruner \cite{ma2023llmpruner}, as references to provide a more comprehensive comparison for our neuron partition method. Neuron Partition not only demonstrates superior performance but also eliminates the dependence on labeled data and explicit gradient computation, thereby enabling more efficient deployment of training-free sparsity methods. 

\begin{table}[h]
\centering
\small
\caption{Comparison between neuron partition with LLM-Pruner.}
\vspace{-6pt}
\resizebox{\linewidth}{!}{
\begin{tabular}{l|c|cccccc|c}
\toprule
\textbf{Model} 
& \textbf{Metrics} & \textbf{Single Obj.} & \textbf{Two Obj.} & \textbf{Counting} & \textbf{Colors} & \textbf{Position} & \textbf{Color Attri.} & \textbf{Overall}$\uparrow$ \\
\toprule
\multirow{2}{*}{{Ming-Omni}}  
& LLM-Pruner & 0.96 & 0.82 & 0.72 & 0.85 & 0.47 & 0.55 & \underline{0.70}
 \\
 & Neuron Partition & 0.96 & 0.81 & 0.58 & 0.86 & 0.49 & 0.56 & \underline{0.71}  \\
\bottomrule
\end{tabular}}
\label{tab:llmpruner}
\end{table}

On the other hand, neuron partition also achieve competive performance compared to quantization. Specifically, in Table~\ref{tab:vs_quantization}, we apply AWQ \citep{Lin2023AWQAW} to quantize the understanding component of Qwen-Image, i.e., the Qwen-VL backbone. 
Notably, neuron partition achieves an overall score of 0.90 at a 50\% compression ratio, outperforming the 0.88 obtained by 4-bit quantization. This stands in contrast to traditional LLM compression, where pruning at similar ratios typically leads to noticeable performance degradation and remains significantly weaker than 4-bit quantization. 

\begin{table}[h]
\centering
\resizebox{\linewidth}{!}{
\begin{tabular}{l|cccccc|c}
\toprule
\textbf{Model} & \textbf{Single Obj.} & \textbf{Two Obj.} & \textbf{Counting} & \textbf{Colors} & \textbf{Position} & \textbf{Color Attri.} & \textbf{Overall}$\uparrow$ \\
\midrule
Qwen-Image 
& 0.99 & 0.98 & 0.91 & 0.94 & 0.80 & 0.89 & \textbf{0.92} \\
w/ 50\% neuron partition 
& 0.99 & 0.94 & 0.94 & 0.93 & 0.76 & 0.87 & \textbf{0.90} \\
w/ 4-bit quantization 
& 0.99 & 0.97 & 0.93 & 0.92 & 0.79 & 0.70 & \textbf{0.88} \\
\bottomrule
\end{tabular}}
\caption{Comparison between neuron partition and quantization, with AWQ \citep{Lin2023AWQAW} applied to quantize the understanding component. }
\label{tab:vs_quantization}
\end{table}

\section{Calibration Data for Neuron Partitioning}

\begin{figure}
  \centering
\vspace{-50pt}
\includegraphics[width=\linewidth]{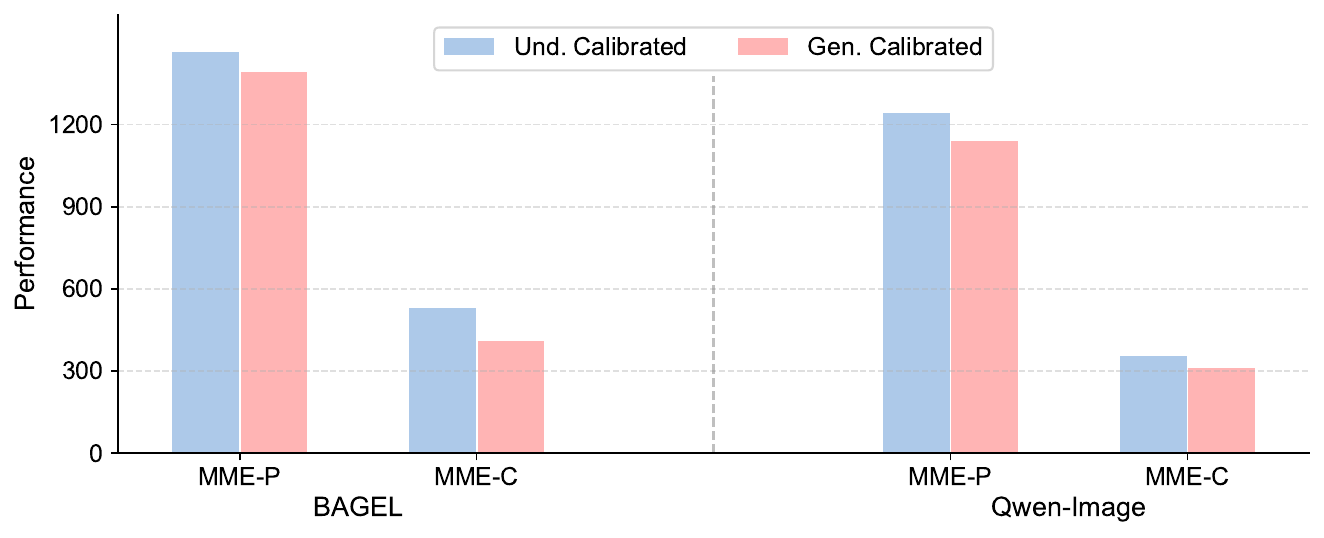} 
\vspace{-20pt}
\caption{Ablation on the choice of calibration datasets for understanding tasks.}   
\vspace{-10pt}
\label{fig:ablation_und}
\end{figure}

\begin{figure}
  \centering
\vspace{-50pt}
\includegraphics[width=\linewidth]{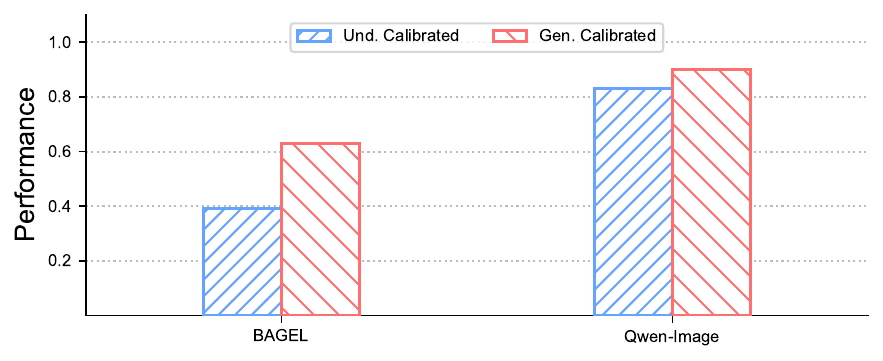} 
\vspace{-20pt}
\caption{Ablation of calibration datasets on generation tasks.}  
\label{fig:ablation_gen}
\end{figure}

The choice of calibration data influences the measurement of activation scores and thus determines which neurons are retained. To comprehensively assess this effect, we conduct an ablation study on MME for understanding and GenEval for generation. Specifically, we sample a small number of instances from each dataset for calibration, compress the models using these calibration sets, and then evaluate them on MME and GenEval, as shown in Figures~\ref{fig:ablation_und} and \ref{fig:ablation_gen}, respectively. Notably, calibration data that align with the target task consistently yield better performance in both understanding and generation. This suggests that unified models rely on distinct neuron partitions for different tasks. Moreover, our test-time few-shot compression, which directly uses a few test samples for calibration, seamlessly adapts to downstream tasks and achieves competitive performance. 



\end{document}